%% file: main.tex
\documentclass[runningheads]{llncs}

\usepackage{eccv/eccv}

\usepackage{eccv/eccvabbrv}

\usepackage{graphicx}
\usepackage{booktabs}
\usepackage{amsmath}
\usepackage{multirow}
\usepackage[table]{xcolor}

\usepackage{tikz}
\usetikzlibrary{arrows.meta,shapes.geometric}

\usepackage[accsupp]{axessibility}  % Improves PDF readability for those with disabilities.

\usepackage{hyperref}

\usepackage{orcidlink}

\input{macros}

\begin{document}

% ---------------------------------------------------------------
\title{GlobalSplat: Efficient Feed-Forward 3D Gaussian Splatting via Global Scene Tokens}

% TODO REVIEW: If the paper title is too long for the running head, you can set
% an abbreviated paper title here. If not, comment out.
\titlerunning{GlobalSplat}

% TODO FINAL: Replace with your author list. 
% Include the authors' OCRID for the camera-ready version, if at all possible.
\author{Roni Itkin\inst{1} \and
Noam Issachar\inst{1} \and
Yehonatan Keypur\inst{1} \and Xingyu Chen \inst{2}
\and \\ Anpei Chen \inst{2}
\and Sagie Benaim \inst{1}
}

% TODO FINAL: Replace with an abbreviated list of authors.
\authorrunning{R.~Itkin et al.}
% First names are abbreviated in the running head.
% If there are more than two authors, 'et al.' is used.

\institute{The Hebrew University of Jerusalem \and
Westlake University }

\maketitle

\newcommand{\fix}{\marginpar{FIX}}
\newcommand{\new}{\marginpar{NEW}}

\input{00_abstract}
\input{01_introduction}
\input{02_related_work}
\input{03_method}

\input{04_experiments}
\input{05_conclusion}

\clearpage

\section*{Acknowledgments}
We acknowledge EuroHPC JU for awarding the project ID EHPC-AIF-2025SC02-060 access to Leonardo at CINECA, Italy. This research was also supported by The Israel Science Foundation (grant No. 2416/25).

\bibliography{references}
\bibliographystyle{eccv/splncs04}

\clearpage

\input{06_appendix}

\end{document}

%% file: macros.tex
\newif\ifdraft
\drafttrue
\ifdraft
\newcommand{\roni}[1]{{\color{magenta}[\textbf{Roni:} #1]}}
\newcommand{\noam}[1]{{\color{cyan}[\textbf{Noam:} #1]}}
\newcommand{\yonatan}[1]{{\color{blue}[\textbf{Yonatan:} #1]}}
\newcommand{\xingyu}[1]{{\color{purple}[\textbf{Xingyu:} #1]}}
\newcommand{\anpei}[1]{{\color{red}[\textbf{Anpei:} #1]}}
\newcommand{\sagie}[1]{{\color{green}[\textbf{Sagie:} #1]}}

\else
\newcommand{\roni}[1]{}
\newcommand{\noam}[1]{}
\newcommand{\yonatan}[1]{}
\newcommand{\xingyu}[1]{}
\newcommand{\anpei}[1]{}
\newcommand{\sagie}[1]{}

\fi

\newcommand{\methodname}{GlobalSplat}

%% file: 00_abstract.tex
\begin{abstract}
The efficient spatial allocation of primitives serves as the foundation of 3D Gaussian Splatting, as it directly dictates the synergy between representation compactness, reconstruction speed, and rendering fidelity. Previous solutions, whether based on iterative optimization or feed-forward inference, suffer from significant trade-offs between these goals, mainly due to the reliance on local, heuristic-driven allocation strategies that lack global scene awareness. Specifically, current feed-forward methods are largely pixel-aligned or voxel-aligned. By unprojecting pixels into dense, view-aligned primitives, they bake redundancy into the 3D asset. As more input views are added, the representation size increases and global consistency becomes fragile.
To this end, we introduce \methodname{}, a framework built on the principle of \emph{align first, decode later}. Our approach learns a compact, global, latent scene representation that encodes multi-view input and resolves cross-view correspondences before decoding any explicit 3D geometry. Crucially, this formulation enables compact, globally consistent reconstructions without relying on pretrained pixel-prediction backbones or reusing latent features from dense baselines. Utilizing a coarse-to-fine training curriculum that gradually increases decoded capacity, \methodname{} natively prevents representation bloat. On RealEstate10K and ACID, 
our model achieves competitive novel-view synthesis performance while utilizing as few as 16K Gaussians, significantly less than required by dense pipelines, obtaining a light 4MB footprint. Further, \methodname{} enables significantly faster inference than the baselines, operating under 78 milliseconds in a single forward pass.
Project page is available at \url{https://r-itk.github.io/globalsplat/}.

\end{abstract}

%% file: 01_introduction.tex
\section{Introduction}
\label{sec:intro}

The efficient spatial allocation of primitives serves as the foundation of 3D Gaussian Splatting (3DGS), directly dictating the synergy between representation compactness, reconstruction speed, and rendering fidelity. We tackle this allocation challenge within the setting of feed-forward novel view synthesis (NVS). We aim to generate a highly compact set of 3D Gaussians from multiple input views in a single network pass, utilizing a fast and lightweight architecture. Crucially, as the number of input views increases to provide broader scene coverage, the network must effectively consolidate this multi-view information simultaneously. It must do so while maintaining a strictly compact 3D representation, utilizing a constant number of scene tokens, independent of the input size.

\begin{figure}[!t]
    \centering
    
    \begin{minipage}{\textwidth}
        \centering
        \small Baseline Per-Pixel Approaches \\ 
        \vspace{1pt}
        \includegraphics[width=0.97\textwidth,clip]{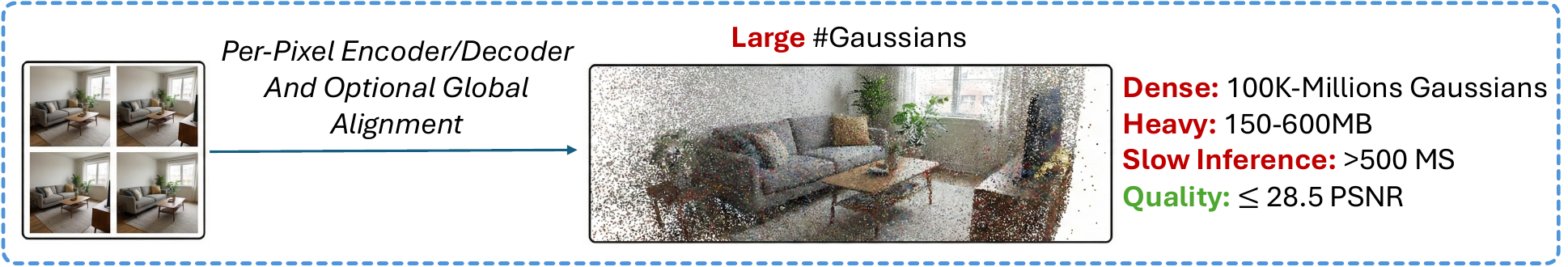} \\ 
        \vspace{6pt}
        \small \methodname{} (Ours) \\  
        \vspace{1pt}
        \includegraphics[width=0.97\textwidth,clip]{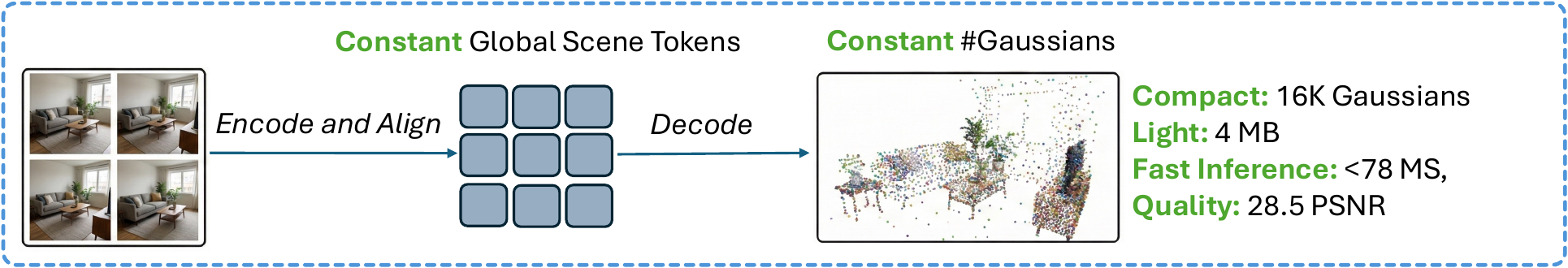}
    \end{minipage}

    \vspace{8pt}
    \makebox[\textwidth]{\rule{0.97\textwidth}{0.4pt}}
    \vspace{8pt}

    \begin{minipage}{\textwidth}
        \centering
        \begin{minipage}{0.62\textwidth}
            \centering
            \includegraphics[width=0.48\linewidth,trim=20pt 0 0 0,clip]{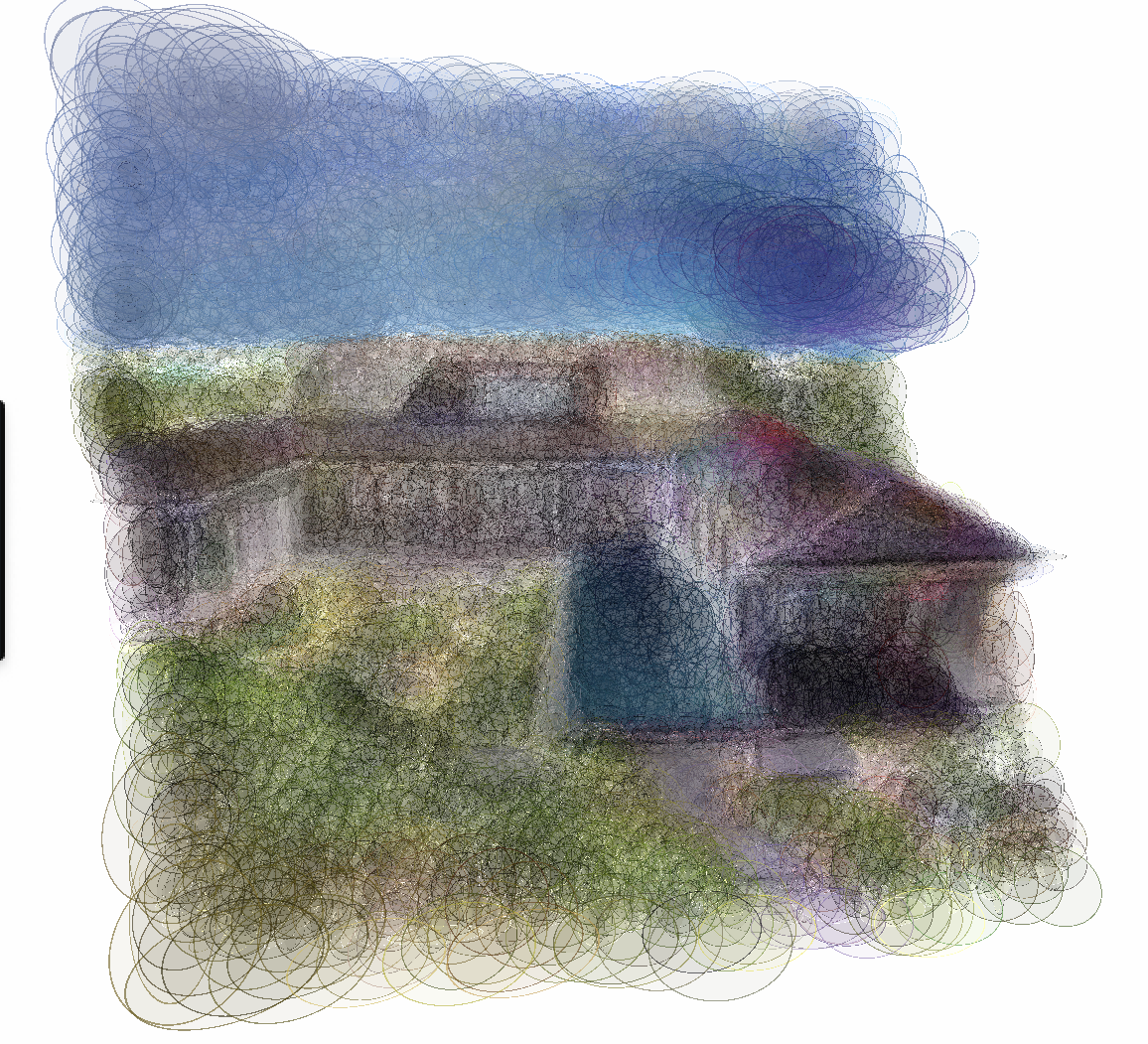}
            \hfill
            \includegraphics[width=0.48\linewidth]{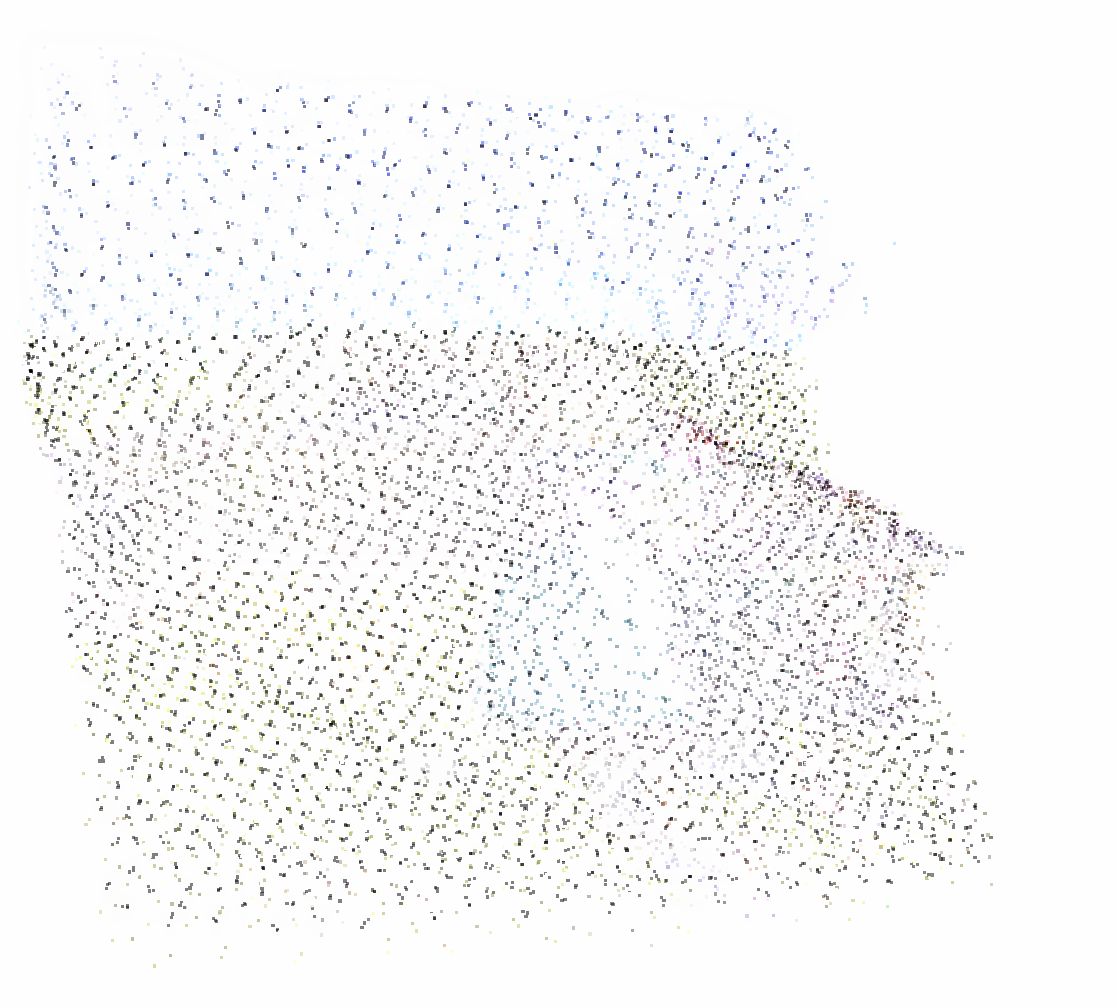}
        \end{minipage}
        \hfill
        \vrule width 0.4pt
        \hfill
        \begin{minipage}{0.33\textwidth}
            \centering
            \includegraphics[width=\linewidth]{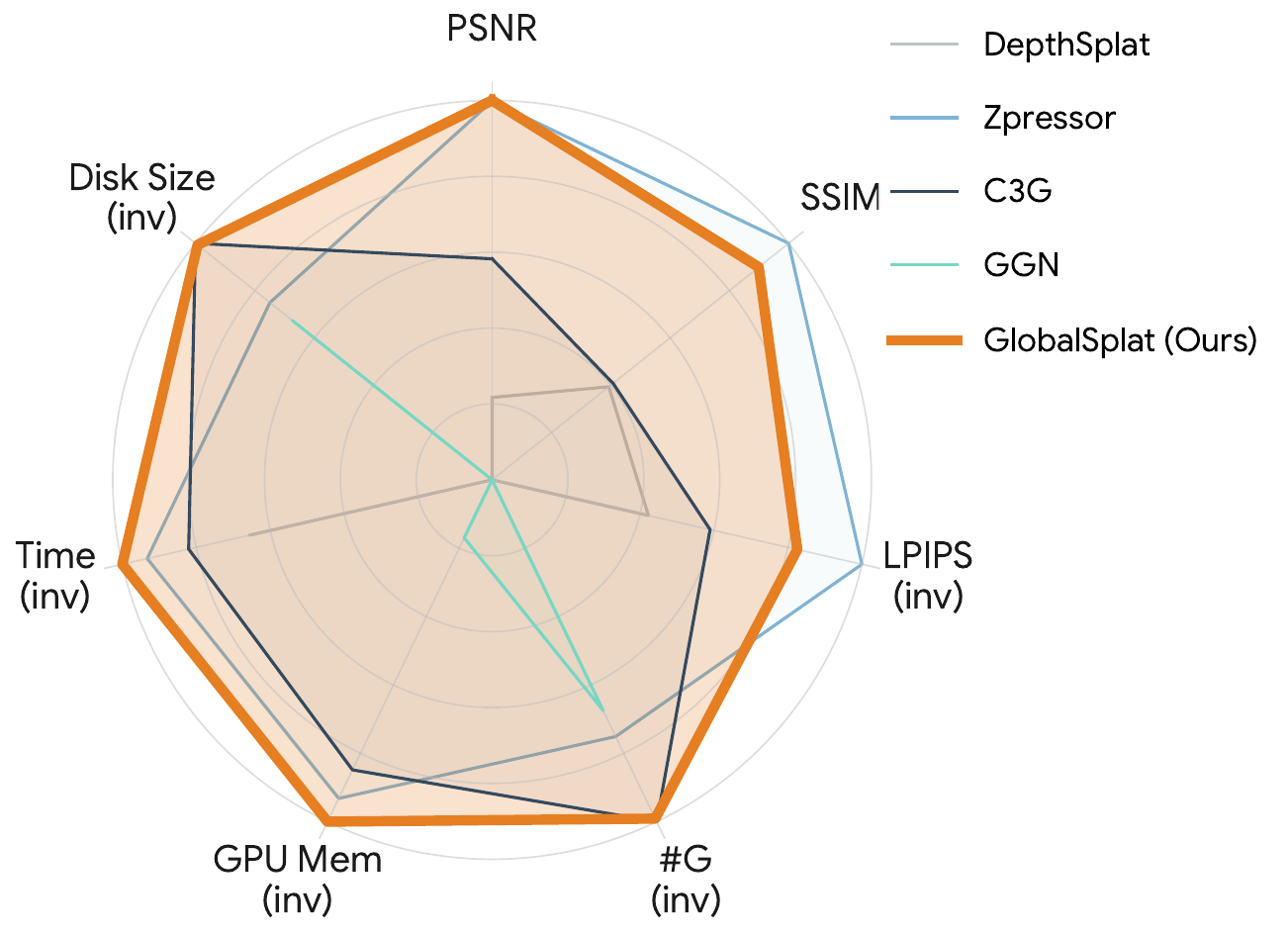}
        \end{minipage}
    \end{minipage}

    \caption{\textbf{Align First, Decode Later.} \textbf{Top:} Existing feed-forward 3D Gaussian Splatting pipelines rely on view-centric, per-pixel primitive allocation. As the number of input views increases, these approaches bake massive redundancy into the 3D representation, scaling to hundreds of thousands or millions of Gaussians. In contrast, \textit{\methodname{}} aggregates multi-view inputs into a fixed set of global latent scene tokens before decoding geometry. This achieves an optimal balance: for instance, for 24 input views on RealEstate10K, \methodname{} delivers highly competitive novel-view synthesis quality (28.5 PSNR), utilizing only 16K Gaussians and offering ultra-low GPU memory usage (1.79 GB), minimal disk size ($<$4 MB), and extremely fast inference times ($<$78 ms).
    \textbf{Bottom Right:} This scene-centric approach translates to a significantly stronger practical operating point in comparison to state-of-the art approaches, as demonstrated in the radar chart (evaluated on 24 input views on RealEstate10K).
    \textbf{Bottom Left:} By decoding from a global scene context (without a fixed grid), \methodname{} stays sparse and places primitives at occupied 3D locations. This yields an adaptive allocation where low-frequency regions are covered by fewer Gaussians with larger spatial support (higher coverage), enabling complex scenes with far fewer primitives.
    We visualize the Gaussians as disks corresponding to their scale, as well as the Gaussian centers using a point cloud. As can be seen, this allows the model to effectively capture complex environments with drastically fewer primitives. }
    \vspace{-0.5cm}
    \label{fig:teaser}
\end{figure}

Recently, feed-forward 3DGS approaches~\cite{charatan2024pixelsplat, zhang2024gs, chen2024mvsplat, wang2024freesplat, jiang2025anysplat, ye2025yonosplat, long2026idesplat, zhang2024gaussian, bai2025graphsplat, wang2025volsplat, wang2025zpressor, song2026tinysplat, an2025c3g} 
have made strong progress on generalizable NVS by predicting explicit Gaussian scene representations directly from input views. 
This is a compelling direction because, once predicted, the output 3DGS asset can be rendered efficiently to target novel views without additional per-scene optimization or per-view generation as in diffusion-based video generation methods. 
However, most existing feed-forward 3DGS pipelines decode scene primitives from dense, view-aligned intermediates (e.g., pixel-aligned predictions, lifted per-view depth/features, or voxel-aligned outputs), attempting to reconcile them globally only afterward. 
This design ties primitive formation to hand-crafted local image or grid structures rather than the scene's intrinsic global structure, and pushes global consistency to a late stage.
Consequently, as the number of input views grows, the model must merge increasingly many view-anchored predictions. This inherently bakes redundancy into the representation and makes large-context reconstruction harder to scale robustly. 
In practice, this leads to an unfavorable trade-off: adding more views improves scene coverage, but simultaneously inflates the representation size and makes reconstruction quality less stable across regions of dense overlap, as illustrated in the top row of Fig.~\ref{fig:teaser}.

Our work, \methodname{}, operates on an opposite appraoch: 
\emph{align first, decode later}, as illustrated in the second row of Fig.~\ref{fig:teaser}.
Specifically, instead of forming Gaussians from per-view dense outputs, \methodname{} first fuses all input views into a globally aligned latent scene representation, and only then decodes an explicit set of 3D Gaussians. 
This shifts reconstruction from view-centric primitive formation to scene-centric primitive formation.
As a result, primitive allocation is driven by scene structure rather than image-grid support, enabling more efficient Gaussian placement while improving global coherence.
The output remains a standard explicit 3DGS representation, preserving the rendering and deployment benefits of Gaussian splatting. 

To achieve this, \methodname{} employs a dual-branch iterative attention architecture that disentangles geometry and appearance, fusing features from all input views into a fixed number of latent scene tokens. A specialized decoder then transforms these globally-aware tokens into explicit 3D Gaussians. By integrating a coarse-to-fine capacity curriculum during training, we naturally prevent the representation from bloating.

We evaluate \methodname{} on RealEstate10K and ACID in 
large-context settings (e.g., 16--36 input views) 
against strong feed-forward 3DGS baselines.
Our results demonstrate a significantly stronger practical operating point for feed-forward reconstruction. \methodname{} maintains a strict, view-invariant budget of 16K Gaussians (and occupying only 4 MB), regardless of the number of input views. It achieves highly competitive novel-view synthesis quality (e.g., 28.5 PSNR for 24 views on RealEstate10K) while utilizing a fraction of the GPU memory during inference (1.79 GB peak memory) and encoding scenes in just under 78 ms. 
To summarize, we provide the following contributions: 
\begin{itemize}
    \item We identify a key limitation of current feed-forward 3DGS pipelines in large-context settings: primitives are typically formed in dense view-aligned spaces and only aligned globally afterward. 
    This initial view-centric primitive formation produces an excessive number of Gaussians, which becomes the primary scalability bottleneck as more views are added.
    \item We propose \methodname{}, a feed-forward 3DGS framework that first builds a globally aligned latent scene representation and then decodes explicit 3D Gaussians (\emph{Align First, Decode Later}).
    \item 
    We demonstrate that this yields a significantly stronger operating point for feed-forward reconstruction. By maintaining an ultra-compact 2K-32K Gaussian representation ($<$4 MB), \methodname{} reduces the number of primitives by over 99\% compared to dense baselines. Furthermore, it requires only 1.79 GB of peak GPU memory and generates scenes in under 78 milliseconds.

\end{itemize}

%% file: 02_related_work.tex
\section{Related Work}
\label{sec:related_work}

\paragraph{Optimization-Based Novel-View Synthesis.}
Optimization-based neural rendering methods produce high-quality per-scene novel-view synthesis (NVS). NeRF and its extensions represent scenes as implicit radiance fields~\cite{mildenhall2021nerf}, while 3D Gaussian Splatting (3DGS) introduced an explicit alternative, representing a scene as a set of anisotropic Gaussians that can be rendered efficiently~\cite{kerbl2023gaussian}. While 3DGS provides an explicit representation of the scene, it typically does so with a significant number of Gaussians. A series of compression-oriented methods reduces storage by 25--100$\times$ or more via quantization, entropy modeling, masking, and learned codebooks over Gaussian attributes, while largely preserving visual fidelity~\cite{niedermayr2024compressed, chen2025hacpp, wang2024contextgs, lee2024compact}. Structural approaches such as ProtoGS and GoDe cut complexity by learning Gaussian prototypes and hierarchical levels of detail, demonstrating that many raw Gaussians can be expressed through a smaller shared or layered set without degrading quality~\cite{gao2025protogs,disario2025gode}. These methods target strong scene-specific reconstruction, but still rely on per-scene optimization.

\paragraph{Feed-Forward 3D Reconstruction.} A major step in feed-forward 3D reconstruction is DUSt3R~\cite{dust3r}, which directly predicts pixel-aligned pointmaps from image pairs without per-scene optimization. Subsequent methods~\cite{Fast3R,VGGT,mapanything,da3} extend this paradigm to multi-view settings with large-scale global attention. While effective, full global attention causes memory and computation to grow rapidly with input length. To improve large-context scalability, recent streaming methods~\cite{StreamVGGT,spann3r,MUSt3R} introduce memory mechanisms for incremental reconstruction. CUT3R~\cite{cut3r} adopts a persistent recurrent state for continuous 3D perception, and TTT3R~\cite{ttt3r} reformulates online updates from a test-time training perspective to better mitigate forgetting over long sequences. These advances motivate scene-level global aggregation in feed-forward pipelines. Our goal is complementary: decoding a compact explicit 3DGS asset for efficient downstream NVS.

\paragraph{Feed-Forward Novel-View Synthesis.}
To avoid per-scene optimization, generalizable feed-forward NVS predicts scene representations in a single forward pass. 
Early learning-based methods tackled this by predicting discrete or layered proxy geometries. Seminal approaches utilized plane-sweep volumes \cite{kalantari2016learning}, Multi-Plane Images (MPIs) such as Stereo Magnification \cite{zhou2018stereo}, Local Light Field Fusion (LLFF) \cite{mildenhall2019llff}, and DeepView \cite{flynn2019deepview}, as well as feature point clouds like SynSin \cite{wiles2020synsin} to warp and blend source views into novel perspectives. While these representations enable fast synthesis, they often struggle with large baseline changes and complex occlusions. Subsequent methods shifted towards implicit continuous fields (e.g., PixelNeRF~\cite{yu2021pixelnerf}, IBRNet~\cite{wang2021ibrnet}, MVSNeRF~\cite{xu2024murf}, and MuRF~\cite{xu2024murf}). These methods improve amortization but remain costly at render time. 

Feed-forward 3DGS methods then emerged, including pixelSplat~\cite{charatan2024pixelsplat}, GS-LRM~\cite{zhang2024gs}, MVSplat~\cite{chen2024mvsplat}, and FreeSplat~\cite{wang2024freesplat}. These methods commonly rely on dense pixel- or view-aligned intermediates, whose memory and compute overhead tends to grow with input-view count. Later work improves geometric robustness~\cite{long2026idesplat}, introduces Gaussian-level aggregation via graph interaction and pooling~\cite{zhang2024gaussian, bai2025graphsplat}, moves to voxel-aligned prediction~\cite{wang2025volsplat}, or supports uncalibrated inputs with joint pose and Gaussian estimation~\cite{jiang2025anysplat, ye2025yonosplat}. Several recent methods focus on scalability and compactness. ZPressor~\cite{wang2025zpressor} and TinySplat~\cite{song2026tinysplat} compress view features or predicted Gaussians, but still rely on view-centric intermediates during prediction. Recently, LVSM~\cite{jin2024lvsm} proposed encoding all input views into a fixed set of latent tokens and decoding target views directly from this latent without explicit 3D structure (such as 3DGS). This demonstrates the effectiveness of a single global latent as the primary fusion space, but rendering new views still requires running a heavy decoder network rather than reusing an explicit asset. A concurrent work to ours, C3G~\cite{an2025c3g} aggregates multi-view features with learnable queries to produce a compact set of Gaussians. However, C3G relies on full self-attention and single-Gaussian decoding; in contrast, our approach introduces an iterative, disentangled dual-branch architecture and a coarse-to-fine capacity curriculum. In our experiments, this design provides a stronger large-context quality-efficiency trade-off.

%% file: 03_method.tex
\section{Method}
\label{sec:method}

\input{figures/overview}

An illustration of our method is provided in Fig.~\ref{fig:overview}. In Sec.~\ref{sec:background}, we detail 3D Gaussian Splatting, our explicit output representation. We then describe our method. First, we normalize the scene and extract ray-augmented patch features, as detailed in Sec.~\ref{sec:camera}. Then, in Sec.~\ref{sec:architecture}, we iteratively refine the latent scene tokens via a dual-branch attention architecture before decoding them directly into explicit 3D Gaussians. Finally, in Sec.~\ref{sec:coarse_to_fine} and Sec.~\ref{sec:objective}, we apply a coarse-to-fine capacity curriculum and consistency objectives to progressively refine local details while strictly preventing representation bloat. Training and implementation details are provided in Appendix~\ref{supp:impl}.

\subsection{Preliminaries}
\label{sec:background}

\begin{equation}
    G(x) = \exp\left(-\frac{1}{2}(x - \mu)^T \Sigma^{-1} (x - \mu)\right)
\end{equation}
Additionally, each Gaussian is assigned an opacity value $\alpha \in [0, 1]$ and view-dependent color coefficients $c$ modeled via Spherical Harmonics (SH).

\paragraph{Rendering.} To synthesize an image, the 3D Gaussians are projected onto the 2D image plane. The final pixel color $C$ is accumulated through front-to-back $\alpha$-blending of $M$ overlapping splats sorted by depth:
\begin{equation}
    C = \sum_{i=1}^{M} c_i \alpha'_i \prod_{j=1}^{i-1} (1 - \alpha'_j)
\end{equation}
where $\alpha'_i$ is the product of the base opacity $\alpha_i$ and the projected 2D Gaussian density. Our model treats the scene as a continuous latent volume during the encoding phase before ``splatting'' tokens into this explicit 3D space.

\subsection{Scene Normalization and Input Preparation}
\label{sec:camera}
\subsubsection{Camera Preprocessing}
Following previous work~\cite{jin2024lvsm, ye2025yonosplat}, we map each scene into a canonical coordinate system using a similarity transform so that camera poses have consistent orientation, translation, and scale across scenes.

\paragraph{Canonical frame.}
Given $C$ poses $\{T_i\}_{i=1}^{C}$, we compute an ``average camera'' frame whose origin is the mean camera center
$
\bar{o}=\frac{1}{C}\sum_{i=1}^{C} o_i,
$
and whose axes are obtained by averaging camera viewing directions and re-orthonormalizing.
Let $T_{\text{avg}}$ denote the resulting average pose. We express all cameras in this frame via
\begin{equation}
\hat{T}_i = T_{\text{avg}}^{-1} T_i .
\end{equation}

\paragraph{Scale normalization.}
Let $\hat{o}_i \in \mathbb{R}^3$ be the camera centers after the above alignment (i.e., the translation component of $\hat{T}_i$).
We follow YoNoSplat~\cite{ye2025yonosplat} and define the scene scale as the diameter of the camera constellation:
\begin{equation}
s \;=\; \max_{a,b}\,\|\hat{o}_a-\hat{o}_b\|_2 .
\end{equation}
We then scale all camera translations by $s$:
\begin{equation}
\tilde{o}_i \;=\; \frac{\hat{o}_i}{s}.
\end{equation}

This ``canonical frustum'' initialization provides a strong geometric prior, allowing the model to focus on refining local structure rather than searching for the global scene location.

\subsubsection{Input Context Construction}
While Plücker rays effectively represent line geometry, they lack focal and translation information; we therefore augment them with a per-view camera code that captures the camera's global context.

For each input view $i$, we extract patchified RGB tokens $u_{i,p}^{\mathrm{rgb}}$.
To inject geometric information, we construct a camera token per patch from two parts: (i) a patchified Pl\"ucker-ray embedding and (ii) a per-view camera code that is broadcast to all patches.

We first compute dense Pl\"ucker-ray features and patchify them to obtain $r_{i,p}$, and map each patch with a learned linear layer:
\begin{equation}
\hat{r}_{i,p} = W_{\mathrm{ray}}\, r_{i,p}.
\end{equation}
In parallel, we form a per-view embedding from the absolute camera center and intrinsics. Let $o_i \in \mathbb{R}^3$ be the camera center and $\kappa_i$ the intrinsics. We encode the camera center with Fourier features and the intrinsics with a small MLP:
\begin{equation}
e_i = W_{\mathrm{proj}}\Big([\mathrm{MLP}_{K}(\phi(\kappa_i));\ \mathrm{PE}(o_i)]\Big),
\end{equation}
where $\phi(\kappa_i)$ uses resolution-normalized intrinsics and $\mathrm{PE}$ denotes Fourier positional encoding. We then add this per-view code to every patch:
\begin{equation}
u_{i,p}^{\mathrm{cam}} = \hat{r}_{i,p} + e_i.
\end{equation}
Finally, we concatenate appearance and camera tokens to form the input context:
\begin{equation}
u_{i,p} = \left[u_{i,p}^{\mathrm{cam}} \,;\, u_{i,p}^{\mathrm{rgb}}\right].
\end{equation}
Our ablations show that explicitly re-injecting absolute camera location and focal information improves performance in large-context settings.
\subsection{\methodname{} Architecture}
\label{sec:architecture}
Our architecture is an encoder-decoder with learnable latent tokens, designed to handle 3D redundancy and prevent gradient conflicts.

\noindent \textbf{Learnable Latent Tokens.}
We initialize a set of $M$ learnable latent tokens $\{l_j\}_{j=1}^M \in \mathbb{R}^{M \times d}$, which serve as the foundation for decoding into Gaussian primitives. Crucially, $M$ is fixed and independent of the number of input frames, ensuring scalability and forcing the model to distill the massive redundancy found in overlapping video views. We also incorporate learnable register tokens to facilitate the capture of global contextual information and prevent local feature over-fitting. In our setting we set $M=2048$ and $d=512$.

\noindent \textbf{Dual-Branch Encoder.}
To prevent ``cheating'', where the model uses texture to mask poor structural predictions, we introduce a \textbf{dual-branch encoder} consisting of $B=4$ blocks. As illustrated in Fig.~\ref{fig:overview}, each block processes the latent tokens through parallel \textit{geometry} and \textit{appearance} streams. Given input patch embeddings $x$, the latent tokens $l$ are projected into stream-specific features:
\begin{equation}
    f_{geo}^{(0)}, f_{app}^{(0)} = \mathrm{Proj}_{geo}(l), \mathrm{Proj}_{app}(l).
\end{equation}
Within each branch, we apply a cross-attention mechanism between the input patches $x$ and the stream features, followed by a $L=2$ self-attention blocks:
\begin{equation}
    f_{i}^{(j)} = \mathrm{SelfAtt}_{i}\left(\mathrm{CrossAtt}_{i}(x, f_{i}^{(j)})\right), \quad i \in \{geo, app\}.
\end{equation}

Finally, the two streams are fused using a 2-layer mixer MLP to update the latent tokens for the subsequent block: $l_{geo}^{(j+1)}, l_{app}^{(j+1)} = \mathrm{MLP}(\mathrm{Concat}(f_{geo}^{(j)}, f_{app}^{(j)}))$. This architectural disentanglement ensures that the appearance and texture features can be processed individually, ensuring structural soundness.

\noindent \textbf{Dual-Branch Decoder}
The decoder transforms the refined latent tokens into the final 3D representation. It employs two specialized linear heads to disentangle the geometric properties (positions, scales, quaternions, opacity) from the texture properties (colors):
\begin{equation}
    G_{geo} = \mathrm{Proj}_{geo}^{dec}(l_{geo}^{(B)}), \quad G_{app} = \mathrm{Proj}_{app}^{dec}(l_{app}^{(B)}).
\end{equation}

\subsection{Coarse-to-Fine Training Curriculum}
\label{sec:coarse_to_fine}
To improve training stability, we introduce a stage-wise capacity curriculum over the $K_s=16$ Gaussian candidates predicted by each latent slot. We begin at a \textit{coarse} stage where all 16 candidates in a slot are merged into a single representative Gaussian ($G=1$). As training progresses, we incrementally increase the capacity ($G \in \{2, 4, 8\}$), allowing the model to refine local details only after global geometry has converged. Our final model uses the capacity $G=8$.

\paragraph{Parameter-Aware Reduction.}
To merge a set of candidate Gaussians $\{g_k\}$ into a single Gaussian $\bar{g}$, we derive importance weights $\pi_k$ from a temperature-scaled softmax. The operator performs a weighted aggregation of attributes as further elaborated in Appendix~\ref{supp:impl}.

\subsection{Training Objective}
\label{sec:objective}
The complete objective is $\mathcal{L} = \lambda_{\mathrm{ren}}\mathcal{L}_{\mathrm{ren}} + \lambda_{\mathrm{con}}\mathcal{L}_{\mathrm{con}} + \lambda_{\mathrm{reg}}\mathcal{L}_{\mathrm{reg}}$.

\paragraph{Rendering Loss.} Given target view $I_t$ and camera $\pi_t$, we optimize:
\begin{equation}
\mathcal{L}_{\mathrm{ren}} = \|I_t-\hat I_t\|_2^2 + \lambda_{\mathrm{perc}}\mathcal{L}_{\mathrm{perc}}(I_t,\hat I_t),
\end{equation}
where $\mathcal{L}_{\mathrm{perc}}$ is a perceptual loss. 

\paragraph{Self-Supervised Consistency.} We partition input views into two subsets $\mathcal{I}_{a}$ and $\mathcal{I}_{b}$. We perform independent forward passes and minimize the distance between the resulting geometry using a stop-gradient ($\text{sg}$) operation:
\begin{equation}
\mathcal{L}_{\mathrm{con}} = \|O(\mathcal{I}_{a}) - \text{sg}(O(\mathcal{I}_{b}))\|_2^2 + \|D(\mathcal{I}_{a}) - \text{sg}(D(\mathcal{I}_{b}))\|_2^2,
\end{equation}
where $O$ and $D$ denote the rendered opacity and depth maps.

\paragraph{Regularization.} To ensure structural integrity, we incorporate:
\begin{equation}
\mathcal{L}_{\mathrm{reg}} = \lambda_{\mathrm{thr}}\mathcal{L}_{\mathrm{thr}} + \lambda_{\mathrm{fru}}\mathcal{L}_{\mathrm{fru}},
\end{equation}
which consists of soft thresholding on features, and a frustum constraint which is further elaborated in Appendix~\ref{supp:impl}.

%% file: figures/overview.tex
\begin{figure*}[t!]
    \centering
    \includegraphics[width=1\textwidth]{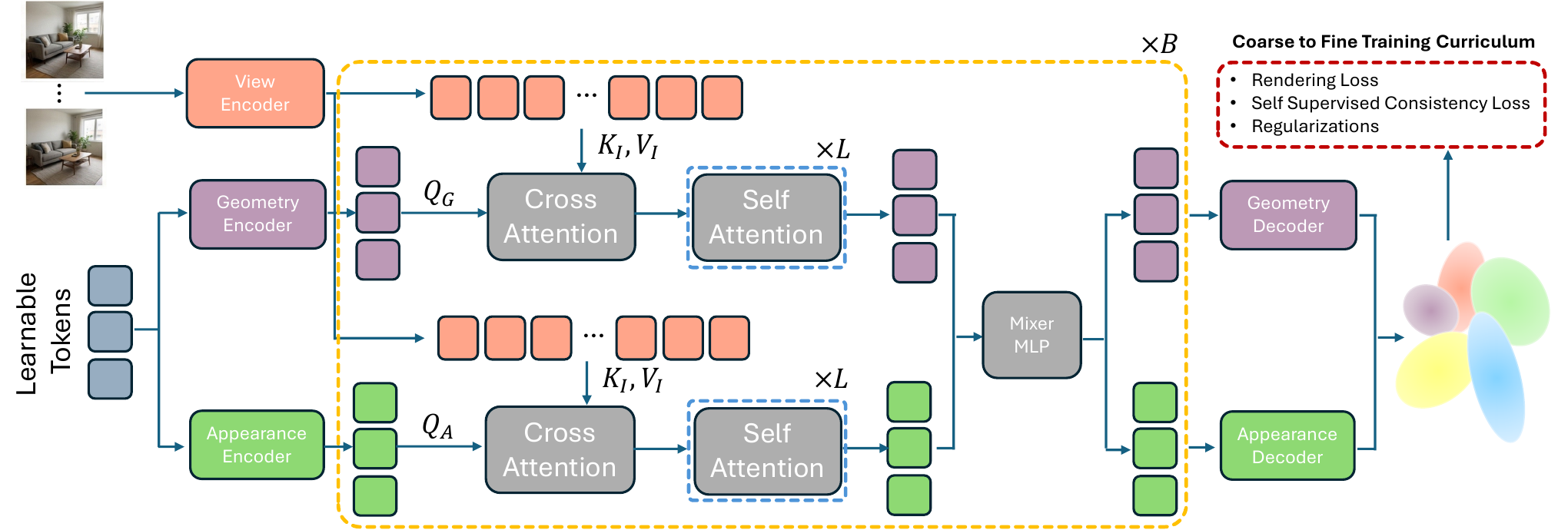}
    \caption{\textbf{\methodname{} Architecture Overview.} Given a sparse set of input views, image features are extracted via a View Encoder. A fixed set of learnable latent scene tokens is iteratively refined through a dual-branch encoder block (repeated $B$ times) designed to explicitly disentangle geometry and appearance. Within each branch, queries ($Q_G, Q_A$) cross-attend to multi-view features ($K_I, V_I$) and self-attend to global context. The streams are fused via a Mixer MLP to update the tokens for the subsequent block. Specialized Geometry and Appearance Decoders then transform these globally-aware tokens into explicit 3D Gaussians. As depicted on the right, the network employs a \textit{Coarse-to-Fine  Training Curriculum} strategy to progressively increase the decoded Gaussian capacity, supervised jointly by rendering, self-supervised consistency, and regularization losses.}
\label{fig:overview}
\end{figure*}

%% file: 04_experiments.tex
\vspace{-0.3cm}
\section{Experiments}
\label{sec:experiments}

\input{figures/real_estate_comparison}

We evaluate our method against state-of-the-art feed-forward NVS baselines. In Sec.~\ref{sec:setup}, we detail the experimental setup, covering the datasets, evaluation protocol, metrics, and baseline methods. In Sec.~\ref{sec:quantitative_eval}, we present a quantitative evaluation of \methodname{} against state-of-the-art feed-forward methods, highlighting our model's compactness, cross-dataset generalization, and computational efficiency. We provide qualitative comparisons in Sec.~\ref{sec:qualitative_eval} to visually demonstrate our reconstruction quality against baseline approaches. Finally, in Sec.~\ref{sec:ablation_study}, we conduct an ablation study to validate our key design choices, including the dual-stream architecture, the coarse-to-fine capacity curriculum, the self-supervised consistency loss, and the explicit injection of camera metadata. Finally, in Sec.~\ref{sec:limitations}, we consider our method's limitations.

\subsection{Experimental Setup}
\label{sec:setup}

\textbf{Datasets.}
We evaluate on \textit{RealEstate10K}~\cite{zhou2018stereo} and \textit{ACID}~\cite{liu2021infinite}, two standard benchmarks for feed-forward and generalizable novel view synthesis. RealEstate10K consists of a large collection of indoor and outdoor real estate video clips, typically featuring forward-facing camera trajectories and room walkthroughs with SLAM-derived camera poses. In contrast, ACID (Aerial Coastline Imagery Dataset) comprises drone and aerial video sequences of natural, unbounded landscapes, equipped with Structure-from-Motion (SfM) camera trajectories.
We use RealEstate10K as our primary training and evaluation benchmark, while ACID serves as a robust zero-shot testbed to evaluate cross-dataset generalization to vastly different, wide-open environments.

\input{figures/comparison_acid}
\input{figures/efficiency}

\noindent \textbf{Protocol.}
We follow the \mbox{C3G}'s~\cite{an2025c3g} RealEstate10K evaluation protocol for both RealEstate10K and ACID, which is built on the standard NoPoSplat~\cite{ye2024no} evaluation split \texttt{assets/evaluation\_index\_re10k.json}. 
The index specifies the anchor \emph{two-view} context and held-out target frames; for multi-view evaluation, we keep the same targets and \emph{inflate} the context by adding extra context frames from the same video segment (i.e., we expand the original two context views to \textit{12}, \textit{24}, and \textit{36} views by sampling additional frames between the anchor context views), following the multiview protocol used by \mbox{C3G}.
All methods are evaluated at \textit{256$\times$256} resolution. % %{index=2}
Given the selected context views, each method predicts a 3D Gaussian scene representation in a feed-forward manner and is evaluated by rendering the held-out target views.

\input{figures/qualitative_eval}

\noindent \noindent \textbf{Metrics.}
We report PSNR, SSIM, and  LPIPS.
For Gaussian-based methods, we additionally report the number of Gaussians, \#G(K), to measure representation compactness.
We also compare efficiency in terms of peak GPU memory, inference time (time of a single forward pass), and size on disk.

\noindent \textbf{Baselines.}
We compare against representative recent methods, including NoPoSplat~\cite{ye2024no}, AnySplat~\cite{jiang2025anysplat}, EcoSplat~\cite{park2025ecosplat}, 
%World Mirror~\cite{jiang2025anysplat}, 
DepthSplat~\cite{xu2025depthsplat}, GGN~\cite{zhang2024gaussian}, Zpressor~\cite{wang2025zpressor}, and the concurrent work of C3G~\cite{an2025c3g}.
We also report results for LVSM~\cite{jin2024lvsm}, which is not a Gaussian-splatting based method.
For \textit{NoPoSplat} and \textit{AnySplat}, we use the numbers reported by the concurrent work \mbox{C3G}. 
For \textit{EcoSplat}, we report the numbers available in the original paper, since code and evaluation outputs were not available to us for re-evaluation under our setup. All other baselines are evaluated using official publicly available code and weights.
We do not report \textit{GGN} results for 36-view evaluation because the publicly available implementation collapses under this setting and fails to produce valid reconstructions.
 For cross-dataset generalization on ACID, we evaluate against DepthSplat~\cite{xu2025depthsplat}, GGN~\cite{zhang2024gaussian}, Zpressor~\cite{wang2025zpressor}, C3G~\cite{an2025c3g}, and LVSM~\cite{jin2024lvsm}. We omit results for NoPoSplat, AnySplat, and EcoSplat on this dataset due to the lack of publicly available code or evaluation outputs required for a consistent comparison under our setup.

\noindent \textbf{Additional results.}
Further quantitative and qualitative results as well as additional ablations, are provided in Appendix~\ref{supp:acid_visual}.

\subsection{Quantitative Evaluation}
\label{sec:quantitative_eval}

Tab.~\ref{tab:main_results} compares \methodname{} with recent feed-forward baselines on RealEstate10K.
Our method achieves strong reconstruction quality while using a \emph{fixed} representation of only \textbf{2K-32K} Gaussians for 12, 24, and 36 input views.
In contrast, several Gaussian-based baselines increase their representation size substantially as the number of input views grows.
This highlights the main advantage of our approach: additional observations do not increase scene complexity.

Among Gaussian-based methods, \methodname{} offers a particularly favorable quality, compactness trade-off.
Compared with the highly compact concurrent work of C3G, it improves image quality by a large margin.
Compared with stronger but much heavier methods such as Zpressor and AnySplat, it uses a dramatically smaller and view-invariant representation.
These results support our central claim that explicit global alignment enables compact yet high-fidelity feed-forward 3D Gaussian reconstruction.

\paragraph{Cross-dataset generalization.}
Tab.~\ref{tab:acid_results} evaluates zero-shot cross-dataset transfer from RealEstate10K to ACID.
\methodname{} remains competitive across all input-view settings, showing that the proposed representation captures transferable scene structure rather than overfitting to the training distribution.
Importantly, this robustness is achieved with the same compact fixed-budget representation used on RealEstate10K.

\begin{table}[t]
\centering
\caption{\textbf{Compactness-Quality trade-off.}
Ablation over the number of latent scene tokens and decoded Gaussians per token on RealEstate10K.
Their product determines the total Gaussian budget, $\#G = (\#\text{Latents}) \times (\text{Splats/Token})$.
Rows are grouped by total Gaussian budget.}
\small
\setlength{\tabcolsep}{5pt}
\begin{tabular}{c c c c c c}
\toprule
\textbf{Total \#G} & \textbf{\#Latents} & \textbf{Splats/Token} & \textbf{PSNR$\uparrow$} & \textbf{SSIM$\uparrow$} & \textbf{LPIPS$\downarrow$} \\
\midrule
2,048   & 256   & 8  & 25.25 & 0.785 & 0.250 \\
2,048   & 2,048 & 1  & 26.83 & 0.838 & 0.198 \\
\midrule
16,384  & 2,048 & 8  & 28.57 & 0.885 & 0.138 \\
\midrule
32,768  & 2,048 & 16 & 28.58 & 0.884 & 0.135 \\
32,768  & 4,096 & 8  & 29.54 & 0.903 & 0.121 \\
\bottomrule
\end{tabular}
\label{tab:compactness_tradeoff}
\end{table}

\paragraph{Efficiency.}
Tab.~\ref{tab:efficiency} shows that the compactness of \methodname{} translates into practical efficiency.
Our method uses the lowest peak GPU memory among the reported methods as well as the fastest inference time while maintaining low footprint on disk. 
These gains are not obtained by sacrificing reconstruction quality; rather, they follow directly from predicting a compact scene representation with a fixed Gaussian budget.
All efficiency benchmarks, including inference latency and peak GPU memory consumption, were measured on a single NVIDIA A100 GPU with 64GB of VRAM ensuring a consistent evaluation environment.

\subsection{Qualitative Evaluation}
\label{sec:qualitative_eval}

Fig.~\ref{fig:qualitative_comparison} provides visual comparisons between \methodname{} and baseline methods across various indoor scenes from the RealEstate10K dataset. Highly compact baselines like C3G struggle to synthesize fine, high-frequency details, often resulting in overly smooth or blurry reconstructions that miss complex textures. 
DepthSplat and GGN tend to introduce structural artifacts and distortions, particularly around object boundaries, thin structures like window blinds, and reflective surfaces. 
While state-of-the-art baselines like Zpressor achieve visual fidelity comparable to ours, they rely on a significantly heavier representation. Specifically, Zpressor requires 393K Gaussians compared to our strict budget of 16K, resulting in higher peak memory usage (3.70 GB vs. 1.79 GB), much slower encoding times (194.20 ms vs. 77.88 ms) and a much more memory heavy representation (134 MB vs. 3.8 MB).
GlobalSplat consistently produces sharp, artifact-free renderings that closely resemble the ground truth. By explicitly disentangling geometry and appearance within a fixed, globally-aligned latent space, our approach effectively recovers intricate room details and maintains robust multi-view consistency without requiring a massive number of primitives.

\subsection{Ablation Study}
\label{sec:ablation_study}

\paragraph{Compactness-Quality Tradeoff.}

Tab.~\ref{tab:compactness_tradeoff} studies how reconstruction quality changes with two factors: the number of latent scene tokens, which controls the latent scene bottleneck, and the number of decoded Gaussians per token, which controls the decoder output density. Their product determines the final Gaussian budget,
\[
\#G = (\#\text{Latents}) \times (\text{Splats/Token}).
\]

The main trend is that increasing latent capacity is substantially more effective than increasing the number of decoded Gaussians per token. At a fixed budget of 2K Gaussians, using 2,048 latents with 1 Gaussian per token clearly outperforms using 256 latents with 8 Gaussians per token. Similarly, at a fixed budget of 32K Gaussians, allocating the budget to more latents is much more effective than allocating it to more Gaussians per token.
In contrast,increasing the decoder density yields only marginal gains. 
Overall, these results indicate that in our feed-forward setting, reconstruction quality is driven primarily by the size of the latent scene representation.

\begin{table}[t]
\centering
\caption{
Model ablation study on RealEstate10K.
For a fair comparison, the single-stream variant increases model width to keep the parameter count comparable to the full model (90M vs.\ 83.4M).
The direct full-capacity variant predicts the full set of Gaussians from the start of training rather than progressively increasing capacity.
The Pl\"ucker-only variant removes the additional camera metadata injected alongside the Pl\"ucker rays.
}
\begin{tabular}{lccc}
\toprule
Variant & PSNR$\uparrow$ & SSIM$\uparrow$ & LPIPS$\downarrow$ \\
\midrule
Ours (full)                    & \textbf{28.57} & \textbf{0.885} & \textbf{0.139} \\
\midrule
Pl\"ucker only                 & 28.30        &  0.880          & 0.140          \\
w/o consistency loss           & 28.15         & 0.876          & 0.143          \\
Single-stream                  & 28.02          & 0.873          & 0.151          \\
Direct full-capacity prediction& 27.69          & 0.867          & 0.150          \\

\bottomrule
\end{tabular}

\label{tab:ablation}
\end{table}
\paragraph{Model Ablation.} Tab.~\ref{tab:ablation} evaluates the main design choices.
Replacing the proposed two-stream design with a single-stream architecture degrades performance even when the single-stream model is widened to keep the parameter count comparable to the full model (90M vs.\ 83.4M).
This indicates that the gain comes from the architectural factorization itself, rather than from model capacity alone.
Removing the coarse-to-fine strategy and predicting the full Gaussian capacity from the beginning of training also reduces performance, showing that progressive capacity growth is important for effective optimization under a compact Gaussian budget.
Finally, removing the additional camera metadata and using only Pl\"ucker rays leads to worse results, indicating that while Pl\"ucker rays provide a strong geometric parameterization, explicitly reinjecting camera metadata remains beneficial.
We also evaluate the impact of our self-supervised consistency objective ($w/o$ consistency loss). Removing this loss leads to a noticeable drop in novel-view synthesis quality and an increase in structural artifacts. 
Overall, the full model consistently performs best, validating the contribution of each component. 

\subsection{Limitations}
\label{sec:limitations}
While \methodname{} establishes a highly efficient operating point for feed-forward 3DGS, it is not without limitations. First, our current architecture relies on a strictly fixed budget of 16K Gaussians. While this is highly effective for room-scale environments (RE10K) and localized aerial trajectories (ACID), unbounded or city-scale environments may eventually exceed the representational capacity of a fixed token set. Future work could explore adaptive or hierarchical token allocation, dynamically scaling the bottleneck based on scene complexity. Second, \methodname{} currently assumes static environments. Extending the global scene tokens to capture temporal dynamics, perhaps via spatio-temporal cross-attention, presents an exciting avenue for efficient 4D reconstruction. Finally, extreme sparse-view settings (e.g., 2 to 3 images) remain challenging due to the lack of sufficient multi-view parallax to properly resolve the global latent space, a direction one can investigate by integrating stronger monocular depth priors.

%% file: figures/real_estate_comparison.tex
\definecolor{top1}{HTML}{FFCCCC} % Light Red
\definecolor{top2}{HTML}{FFE5CC} % Light Orange
\definecolor{top3}{HTML}{FFFFCC} % Light Yellow

\begin{table*}
\caption{\textbf{Quantitative comparison on RealEstate10K.} We report PSNR ($\uparrow$), SSIM ($\uparrow$), LPIPS ($\downarrow$), and the number of Gaussians $\#G(K)$ ($\downarrow$). Ranking highlights: \colorbox{top1}{1st}, \colorbox{top2}{2nd}, and \colorbox{top3}{3rd}. \textcolor{gray}{Gray} text indicates LVSM is not a Gaussian-based approach. We report the $2K, 16K$ and $32K$ Gaussians variants of \methodname{}. Crucially, our method offers a highly favorable quality-compactness trade-off compared to existing baselines. While highly compact methods like C3G sacrifice significant image quality, and stronger methods like Zpressor and AnySplat rely on massive, memory-heavy representations (over 393K and up to 3.3M Gaussians, respectively), \methodname{} offer competitive results using a fraction of the primitives used by baselines. Zpressor-$X$ represents the number of sampled context views used in Zpressor. EcoSplat has no available code and so could not be evaluated on 12/36 views (24 views provided in their paper).}
\resizebox{\textwidth}{!}{
\begin{tabular}{l | cccc | cccc | cccc}
\toprule
\multirow{2}{*}{Method} & \multicolumn{4}{c|}{12 Views} & \multicolumn{4}{c|}{24 Views} & \multicolumn{4}{c}{36 Views} \\
\cmidrule(lr){2-5} \cmidrule(lr){6-9} \cmidrule(lr){10-13}
& PSNR $\uparrow$ & SSIM $\uparrow$ & LPIPS $\downarrow$ & \#G(K) $\downarrow$
& PSNR $\uparrow$ & SSIM $\uparrow$ & LPIPS $\downarrow$ & \#G(K) $\downarrow$
& PSNR $\uparrow$ & SSIM $\uparrow$ & LPIPS $\downarrow$ & \#G(K) $\downarrow$ \\
\midrule
\color{gray}LVSM(non-GS)~\cite{jin2024lvsm} & \color{gray}28.65 & \color{gray}0.898 & \color{gray}0.095 & \color{gray}-- & \color{gray}27.24 & \color{gray}0.874 & \color{gray}0.112 & \color{gray}-- & \color{gray}26.38 & \color{gray}0.855 & \color{gray}0.126 & \color{gray}-- \\
\midrule
NoPoSplat~\cite{ye2024no} & 21.26 & 0.667 & 0.200 & 602 & 21.24 & 0.664 & 0.200 & 1204 & 21.19 & 0.663 & 0.200 & 1806 \\
AnySplat~\cite{jiang2025anysplat} & 23.06 & 0.807 & 0.215 & 1500 & 24.11 & 0.838 & 0.198 & 2636 &24.20 & 0.842 & 0.192 & 3309 \\
EcoSplat~\cite{park2025ecosplat} & -- & -- & -- & -- & 24.72 & 0.822 & 0.183 & 78 & -- & -- & -- & -- \\
DepthSplat~\cite{xu2025depthsplat} & 21.35 & 0.809 & 0.190 & 786 & 19.66 & 0.743 & 0.239 & 1572 & 18.84 & 0.704 & 0.268 & 2359 \\
GGN~\cite{zhang2024gaussian} & 20.11 & 0.710 & 0.271 & 278 & 18.50 & 0.682 & 0.299 & 385 & 17.76 & 0.664 & 0.311 & 466 \\
Zpressor6~\cite{wang2025zpressor} & \colorbox{top3}{28.46} & \colorbox{top1}{0.910} & \colorbox{top1}{0.098} & 393 & \colorbox{top3}{28.51} & \colorbox{top1}{0.911} & \colorbox{top1}{0.097} & 393 & \colorbox{top2}{28.50} & \colorbox{top1}{0.911} & \colorbox{top1}{0.097} & 393\\
Zpressor3~\cite{wang2025zpressor}& 23.63 & 0.846 & 0.157 & 197 & 23.65 & 0.846 & 0.157 &197 & 23.65 & 0.846 &0.157 &197 \\
C3G~\cite{an2025c3g} & 23.61 & 0.740 & 0.203 & \colorbox{top1}{2} & 23.80 & 0.747 & 0.198 & \colorbox{top1}{2} & 23.81 & 0.747 & 0.199 & \colorbox{top1}{2} \\
\midrule
\textbf{GlobalSplat2K (Ours)} & 26.83 &0.838 & 0.198 & \colorbox{top1}{2} & 26.84 & 0.838 & 0.198 & \colorbox{top1}{2}  & 26.84 &  0.838 & 0.200 & \colorbox{top1}{2} \\
\textbf{GlobalSplat16K (Ours)} & \colorbox{top2}{28.57} & \colorbox{top3}{0.885} & \colorbox{top3}{0.138} & \colorbox{top2}{16} & \colorbox{top2}{28.53} & \colorbox{top3}{0.883} & \colorbox{top3}{0.140} & \colorbox{top2}{16} & \colorbox{top3}{28.45} & \colorbox{top3}{0.880} & \colorbox{top3}{0.144} & \colorbox{top2}{16} \\
\textbf{GlobalSplat32K (Ours)} & \colorbox{top1}{29.54} & \colorbox{top2}{0.903} &\colorbox{top2}{0.121} &\colorbox{top3}{32} & \colorbox{top1}{29.48} & \colorbox{top2}{0.901} &  \colorbox{top2}{0.122} & \colorbox{top3}{32} &\colorbox{top1}{29.39} & \colorbox{top2}{0.899} & \colorbox{top2}{0.126} &  \colorbox{top3}{32} \\
\bottomrule

\end{tabular}
}
\vspace{-0.4cm}
\label{tab:main_results}
\end{table*}

%% file: figures/comparison_acid.tex
\begin{table*}[t]
\centering
\setlength{\fboxsep}{1.5pt}

\caption{\textbf{Cross-dataset generalization on ACID.} We report PSNR ($\uparrow$), SSIM ($\uparrow$), LPIPS ($\downarrow$), and the number of Gaussians $\#G(K)$ ($\downarrow$). Ranking highlights: \colorbox{top1}{1st}, \colorbox{top2}{2nd}, and \colorbox{top3}{3rd} (excluding LVSM). \textcolor{gray}{Gray} text indicates LVSM is not a Gaussian-splatting based approach. \methodname{} generalizes robustly across all input-view settings despite using a compact fixed-size representation. 
}

\resizebox{\textwidth}{!}{
\begin{tabular}{l | cccc | cccc | cccc}
\toprule
\multirow{2}{*}{Method} & \multicolumn{4}{c|}{12 Views} & \multicolumn{4}{c|}{24 Views} & \multicolumn{4}{c}{36 Views} \\
\cmidrule(lr){2-5} \cmidrule(lr){6-9} \cmidrule(lr){10-13}
& PSNR $\uparrow$ & SSIM $\uparrow$ & LPIPS $\downarrow$ & \#G(K) $\downarrow$
& PSNR $\uparrow$ & SSIM $\uparrow$ & LPIPS $\downarrow$ & \#G(K) $\downarrow$
& PSNR $\uparrow$ & SSIM $\uparrow$ & LPIPS $\downarrow$ & \#G(K) $\downarrow$ \\
\midrule
\color{gray}LVSM(non-GS)~\cite{jin2024lvsm} & \color{gray}29.23 & \color{gray}0.849 & \color{gray}0.142 & \color{gray}-- & \color{gray}28.29 & \color{gray}0.826 & \color{gray}0.161 & \color{gray}-- & \color{gray}27.61 & \color{gray}0.807 & \color{gray}0.178 & \color{gray}-- \\
\midrule
DepthSplat~\cite{xu2025depthsplat} & 21.45 & \colorbox{top2}{0.769} & \colorbox{top3}{0.220} & 786 & 20.15 & \colorbox{top2}{0.711} & \colorbox{top3}{0.258} & 1572 & 19.60 & \colorbox{top2}{0.681} & \colorbox{top3}{0.279} & 2359 \\
GGN~\cite{zhang2024gaussian} & 21.99 & 0.686 & 0.295 & \colorbox{top3}{287} & 20.90 & 0.657 & 0.314 & 396 & 20.43 & 0.644 & 0.323 & 475 \\
Zpressor~\cite{wang2025zpressor} & \colorbox{top1}{28.44} & \colorbox{top1}{0.859} & \colorbox{top1}{0.140} & 393 & \colorbox{top1}{28.53} & \colorbox{top1}{0.860} & \colorbox{top1}{0.138} & \colorbox{top3}{393} & \colorbox{top1}{28.45} & \colorbox{top1}{0.859} & \colorbox{top1}{0.139} & \colorbox{top3}{393} \\
C3G~\cite{an2025c3g} & \colorbox{top3}{22.24} & 0.598 & 0.332 & \colorbox{top1}{2} & \colorbox{top3}{22.24} & 0.598 & 0.331 & \colorbox{top1}{2} & \colorbox{top3}{22.20} & 0.598 & 0.333 & \colorbox{top1}{2} \\
\midrule
\textbf{GlobalSplat16K (Ours)} & \colorbox{top2}{28.04} & \colorbox{top3}{0.815} & \colorbox{top2}{0.207} & \colorbox{top2}{16} & \colorbox{top2}{28.03} & \colorbox{top3}{0.813} & \colorbox{top2}{0.208} & \colorbox{top2}{16} & \colorbox{top2}{27.99} & \colorbox{top3}{0.810} & \colorbox{top2}{0.213} & \colorbox{top2}{16} \\
\bottomrule
\end{tabular}
}
\vspace{-0.1cm}
\label{tab:acid_results}
\end{table*}

%% file: figures/efficiency.tex
\definecolor{top1}{HTML}{FFCCCC} % Light Red
\definecolor{top2}{HTML}{FFE5CC} % Light Orange
\definecolor{top3}{HTML}{FFFFCC} % Light Yellow

\begin{table}[t]
\centering
\setlength{\fboxsep}{1.5pt}
\caption{\textbf{Efficiency comparison for 24 input views.} We report Peak Mem (peak GPU allocation), Inference Time, and Size on Disk. Ranking highlights: \colorbox{top1}{1st}, \colorbox{top2}{2nd}, and \colorbox{top3}{3rd} (excluding LVSM). \textcolor{gray}{Gray} text indicates LVSM is not a Gaussian-splatting based approach. \methodname{} is the most memory-efficient method, requiring only 1.79 GB of peak memory. It also achieves the fastest inference time (77.88 ms) and maintains an ultra-light 3.8 MB footprint on disk. 
}
\begin{tabular}{lc c c c c | c}
\toprule
Metric & \color{gray}LVSM~\cite{jin2024lvsm} & DepthSplat~\cite{xu2025depthsplat} & Zpressor~\cite{wang2025zpressor} & C3G~\cite{an2025c3g} & GGN~\cite{zhang2024gaussian} & \textbf{Ours16K} \\
\midrule
Peak Mem (GB) & \color{gray}4.60 & 29.84 & \colorbox{top2}{3.70} & \colorbox{top3}{6.04} & 25.08 & \colorbox{top1}{1.79} \\
Inf. Time (ms) & \color{gray}940.00 & 669.50 & \colorbox{top2}{194.20} & \colorbox{top3}{387.14} & 1800.64 & \colorbox{top1}{77.88} \\
Size on Disk (MB) & \color{gray}-- & 534 & \colorbox{top3}{134} & \colorbox{top1}{0.1} & 174 & \colorbox{top2}{3.8} \\
\bottomrule
\end{tabular}
\label{tab:efficiency}
\vspace{-0.5cm}
\end{table}

%% file: figures/qualitative_eval.tex
\begin{figure*}[t!] 
    \centering

    \newcommand{\imgwidth}{0.16\linewidth} 
    
    \setlength{\tabcolsep}{1pt}       
    \renewcommand{\arraystretch}{0.5} 

    \begin{tabular}{cccccc}
        \small Zpressor &\small DepthSplat & \small GGN & \small C3G & \small Ours & \small GT \\
        \vspace{1pt} \\ 

        \includegraphics[width=\imgwidth]{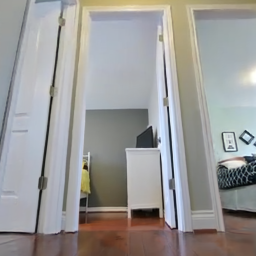} &
        \includegraphics[width=\imgwidth]{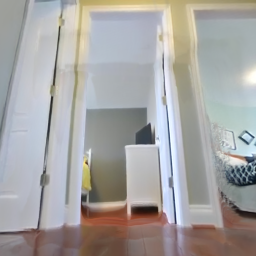} &
        \includegraphics[width=\imgwidth]{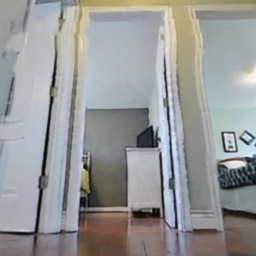} &
        \includegraphics[width=\imgwidth]{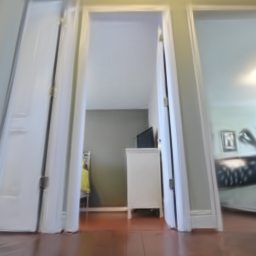} &
        \includegraphics[width=\imgwidth]{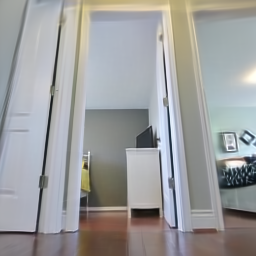} &
        \includegraphics[width=\imgwidth]{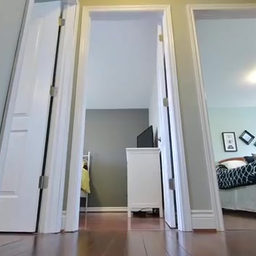} \\

        \includegraphics[width=\imgwidth]{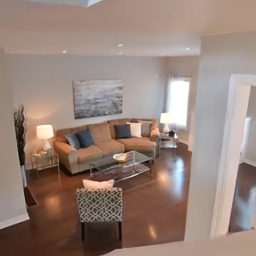} &
        \includegraphics[width=\imgwidth]{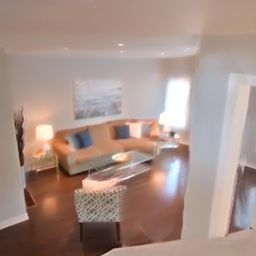} &
        \includegraphics[width=\imgwidth]{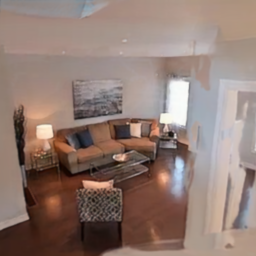} & 
        \includegraphics[width=\imgwidth]{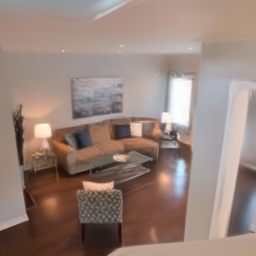} &
        \includegraphics[width=\imgwidth]{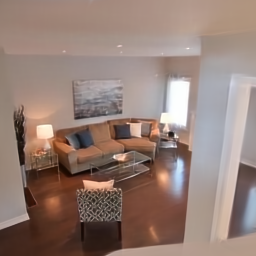}  &
        \includegraphics[width=\imgwidth]{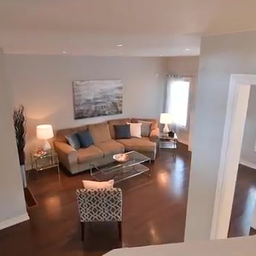} \\

        \includegraphics[width=\imgwidth]{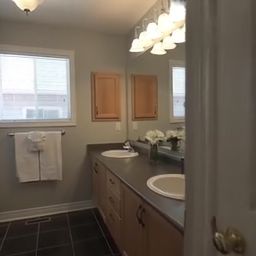} &
        \includegraphics[width=\imgwidth]{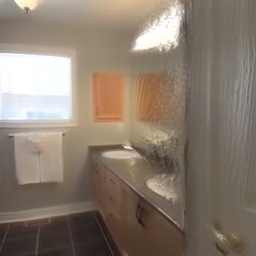} &
        \includegraphics[width=\imgwidth]{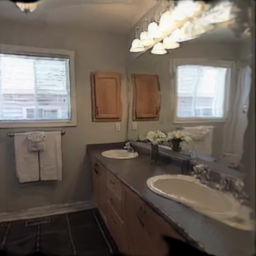} &
        \includegraphics[width=\imgwidth]{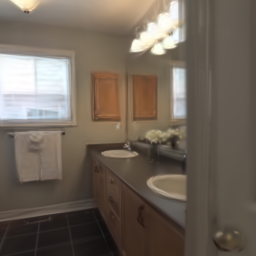} &
        \includegraphics[width=\imgwidth]{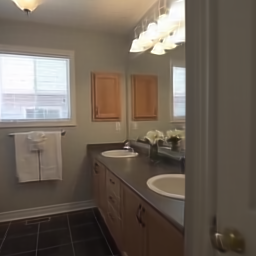} &
        \includegraphics[width=\imgwidth]{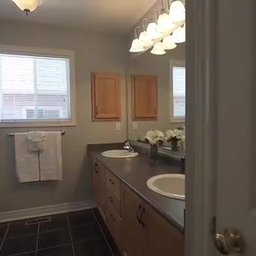} \\

        \includegraphics[width=\imgwidth]{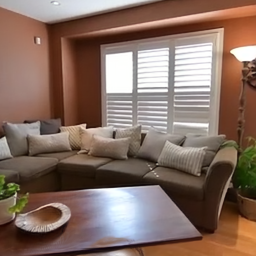} &
        \includegraphics[width=\imgwidth]{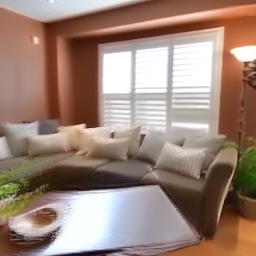} &
        \includegraphics[width=\imgwidth]{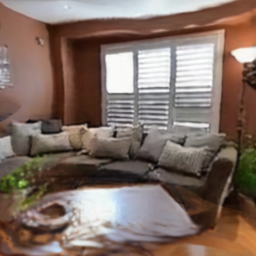} &
        \includegraphics[width=\imgwidth]{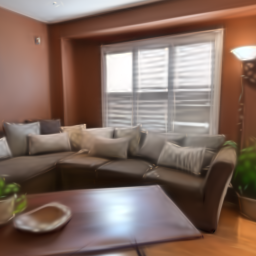} &
        \includegraphics[width=\imgwidth]{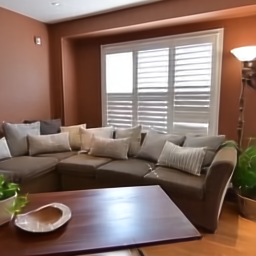} &
        \includegraphics[width=\imgwidth]{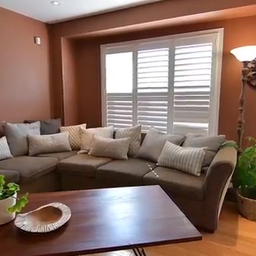} \\

        \includegraphics[width=\imgwidth]{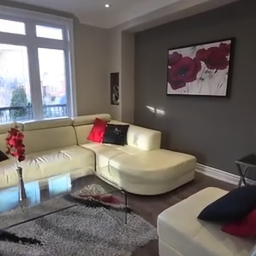} &
        \includegraphics[width=\imgwidth]{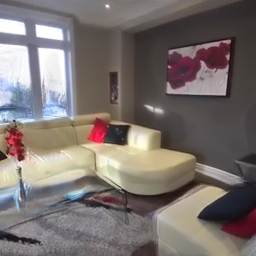} &
        \includegraphics[width=\imgwidth]{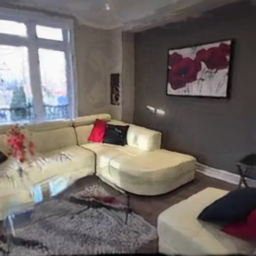} &
        \includegraphics[width=\imgwidth]{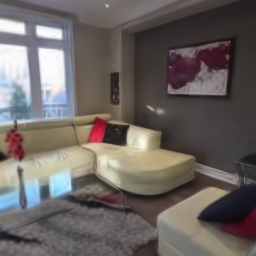} &
        \includegraphics[width=\imgwidth]{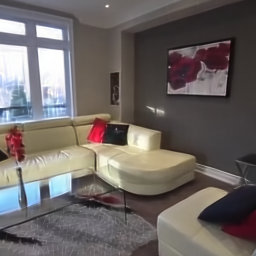} &
        \includegraphics[width=\imgwidth]{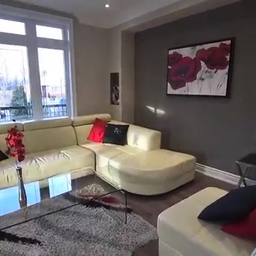} \\

        \includegraphics[width=\imgwidth]{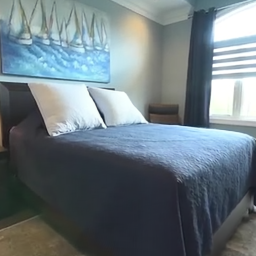} &
        \includegraphics[width=\imgwidth]{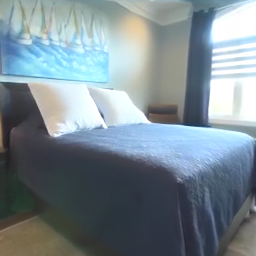} &
        \includegraphics[width=\imgwidth]{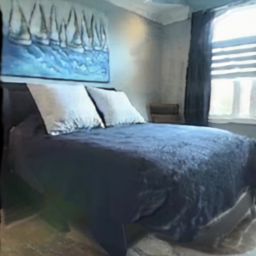} &
        \includegraphics[width=\imgwidth]{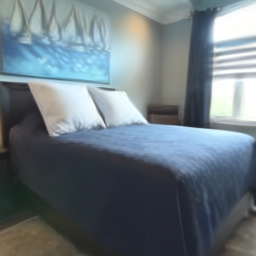} &
        \includegraphics[width=\imgwidth]{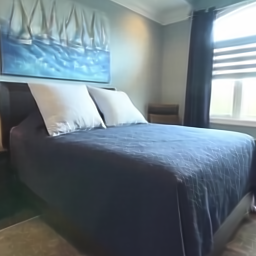} &
        \includegraphics[width=\imgwidth]{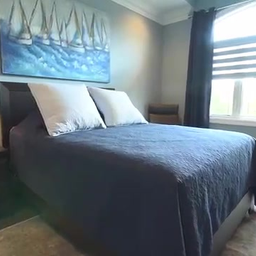} \\

    \end{tabular}
    
    \caption{\textbf{Qualitative comparison}. We compare \methodname{} against baselines (Zpressor, DepthSplat, GGN, C3G) and the ground truth (GT) across 6 different scenes (rows). }
    \label{fig:qualitative_comparison}
    \vspace{-0.5cm}
\end{figure*}

%% file: 05_conclusion.tex
\section{Conclusion}
\label{sec:conclusion}

In this work, we introduced \methodname{}, an efficient feed-forward 3D Gaussian Splatting framework built on the principle of ``align first, decode later''. By aggregating multi-view observations into a compact, fixed-size set of global scene tokens before decoding any explicit 3D geometry, we eliminate the massive redundancy inherent in dense, view-centric pipelines. Equipped with a disentangled dual-branch encoder and a coarse-to-fine training curriculum, \methodname{} achieves highly competitive novel-view synthesis quality on large-context scenes while strictly capping the representation at 2K-32K Gaussians. This ultra-compact $<$4 MB footprint translates to significantly faster inference and generation times, minimal memory utilization, and real-time rendering speeds. Ultimately, \methodname{} establishes a highly practical and scalable operating point for feed-forward 3D scene reconstruction. 

%% file: 06_appendix.tex
\definecolor{top1}{HTML}{FFCCCC} % Light Red
\definecolor{top2}{HTML}{FFE5CC} % Light Orange
\definecolor{top3}{HTML}{FFFFCC} % Light Yellow

\appendix

\section{Qualitative Results on ACID}
\label{supp:acid_visual}
\input{figures/acid_visual}

In addition the supplementary webpage, we provide additional qualitative results on ACID in Fig.~\ref{fig:qualitative_comparison_acid}. 
Overall, the qualitative ranking is consistent with the RealEstate10K results: our reconstructions are visually close to ZPressor and clearly better than the other baselines. In particular, GGN often fails due to errors introduced by merging in the output space, which leads to unstable structure and visible artifacts. DepthSplat tends to produce many redundant splats and depth inaccuracies, and these errors accumulate into noticeably worse renderings. C3G, while highly compact, lacks sufficient expressive capacity to recover fine details and complex scene structure, leading to oversimplified results. In contrast, GlobalSplat produces sharper and more coherent reconstruction showing that our proposed approach transfers well across dataset.

\section{Implementation Details}
\label{supp:impl}

At evaluation time, each image is resized to height 256 while preserving aspect ratio, with the resized width rounded to a multiple of the patch size (8), additionally, the intrinsics camera parameters are updated after resizing. We then apply a centered square crop and, if needed, a final resize to obtain an exact $256\times256$ image. The same deterministic preprocessing is applied to both context and target views.
\subsection{Model Architecture and Decoder}
\label{supp:arch_decoder}

\paragraph{Gaussian parameterization.}
The decoder maps the latent scene tokens to a compact 3D Gaussian representation.
Each Gaussian is parameterized by a 3D mean, anisotropic scale, rotation, opacity, and view-dependent color represented with spherical harmonics (SH).
We use SH degree $3$, corresponding to
\begin{equation}
D_{\mathrm{SH}} = (3+1)^2 = 16
\end{equation}
coefficients per color channel. Rotations are represented using the continuous 6D
parameterization, and the predicted 6D values are modeled as residual offsets from the static rotation.
To bias the initial predictions toward valid visible geometry,
Gaussian means are initialized with a fixed forward offset of 1.5.
Scale is predicted in log space as an offset from -2.
Opacity is predicted in logit space as an offset from -5.

\paragraph{Encoder.}
Our model follows a two-stream encoder--decoder design that separates geometry and appearance processing.
The encoder operates on $8\times 8$ image patches and produces latent scene tokens of dimension $512$. 
The input embedding uses dimension $512$ for RGB features and $256$ for ray features.
We use $2048$ latent tokens and apply $4$ iterative encoder rounds, each with self-attention blocks with depth $2$.

\paragraph{Coarse-to-fine decoding.}
For each latent token, the decoder predicts a fixed set of $16$ Gaussian candidates throughout training.
As introduced in Sec. 3.4 in the main text, rather than progressively introducing new Gaussians, the decoder controls the effective output granularity through a stage-dependent merging mechanism.
At stage $s$, it exposes
\begin{equation}
G = 2^s
\end{equation}
Gaussians per latent token, with $s \in \{0,1,2,3,4\}$.
Thus, stage $0$ corresponds to a single strongly merged Gaussian, while later stages gradually reveal finer structure by reducing the amount of merging.
In the final model, we stop at stage $3$, yielding
\begin{equation}
G = 2^3 = 8
\end{equation}
Gaussians per token.
With $2048$ latent tokens, this gives a final scene representation of
\begin{equation}
N = 2048 \times 8 = 16{,}384
\end{equation}
Gaussians per scene.

\paragraph{Dual-branch prediction.}
The decoder uses two aligned prediction branches.
The geometry branch predicts Gaussian centers, anisotropic log-scales, 6D rotations, opacity logits, and importance score, while the appearance branch predicts spherical harmonics coefficients.
For each latent token $p$, the full set of candidate attributes is
\begin{align}
\mathbf{X}_p &\in \mathbb{R}^{16\times 3}, \\
\mathbf{S}_p &\in \mathbb{R}^{16\times 3}, \\
\mathbf{R}_p &\in \mathbb{R}^{16\times 6}, \\
\mathbf{O}_p &\in \mathbb{R}^{16\times 1}, \\
\mathbf{\ell}_p &\in \mathbb{R}^{16\times 1}, \\
\mathbf{C}_p &\in \mathbb{R}^{16\times (3D_{\mathrm{SH}})} .
\end{align}

\paragraph{Stage-dependent merging.}
At stage $s$, the $16$ candidates are partitioned into $G=2^s$ groups, each of size
\begin{equation}
b_s = \frac{16}{G}.
\end{equation}
A geometry-conditioned gate predicts soft weights within each group.
If $\ell_{p,g,i}$ denotes the gate logit for the $i$-th candidate in group $g$, the normalized gate weight is
\begin{equation}
w_{p,g,i}
=
\frac{\exp(\ell_{p,g,i}/\tau)}
{\sum_{j=1}^{b_s}\exp(\ell_{p,g,j}/\tau)},
\end{equation}
where $\tau$ is a temperature parameter.
For a generic attribute $\mathbf{z}$, the merged output is
\begin{equation}
\bar{\mathbf{z}}_{p,g}
=
\sum_{i=1}^{b_s} w_{p,g,i}\,\mathbf{z}_{p,g,i}.
\end{equation}
Thus, all $16$ candidates are always predicted, but at early stages only a coarse grouped representation is exposed to the renderer.

\paragraph{Smooth stage transition.}
To avoid abrupt transitions between stages, we linearly interpolate between the previous coarser representation and the current less-merged one.
Let $\lambda\in[0,1]$ denote the transition coefficient, for a decoded quantity $\mathbf{z}$,
\begin{equation}
\mathbf{z}^{(s)}
=
(1-\lambda)\,\mathbf{z}^{(s-1)}
+
\lambda\,\bar{\mathbf{z}}^{(s)}.
\end{equation}
This procedure does not introduce new candidates.
Instead, it progressively relaxes the merging applied to the same underlying set of predictions.

\paragraph{Attribute-specific merge rules.}
As introduces in Sec. 3.4 in the main text, positions, rotations, and SH coefficients are merged by weighted averaging. 
Log-scales are merged in log-space with a volume-preserving correction,
\begin{equation}
\log \mathbf{s}^{(s)}
=
\sum_i w_i \log \mathbf{s}_i + \frac{\log b_s}{3},
\end{equation}
and the expanded coarser representation uses the inverse binary split rule
\begin{equation}
\log \mathbf{s}_{\mathrm{child}}
=
\log \mathbf{s}_{\mathrm{parent}} - \frac{\log 2}{3}.
\end{equation}

\noindent Opacity is merged by defining $\alpha_i=\sigma(o_i)$ and $u_i=1-\alpha_i$, we compute
\begin{equation}
\log u_{\mathrm{parent}}
=
\sum_i w_i \log u_i,
\end{equation}
or equivalently,
\begin{equation}
\log(1-\alpha_{\mathrm{parent}})
=
\sum_i w_i \log(1-\alpha_i).
\end{equation}
During expansion, the parent is evenly split:
\begin{equation}
\log u_{\mathrm{child}} = \frac{1}{2}\log u_{\mathrm{parent}}.
\end{equation}

\subsection{Data Sampling and Training Setup}
\label{supp:training_setup}

\paragraph{Crop and resize.}
Training and evaluation use different preprocessing. During training, each view is first cropped around its principal point by taking the largest valid rectangle centered at $(c_x,c_y)$, and the intrinsics are updated accordingly. The cropped image is then resized to the target resolution, with a small additional crop augmentation applied before the final resize/crop to $256\times256$. This augmentation is sampled once per training sample and shared across all views, making it multi-view consistent. Camera intrinsics are updated after every crop and resize operation.

\paragraph{View sampling.}
Training samples are drawn from monocular video sequences.
We first sample a random start frame and define a temporal window whose length is drawn uniformly from $[40,220]$ frames.
Input and target views are then sampled uniformly from this window.
In all experiments, we use $13$ input views and $12$ target views from the same local segment.
This exposes the model to a broad range of camera baselines while ensuring that all views remain geometrically related.

\paragraph{Training configuration.}
All images are resized to $256\times 256$.
We apply color jitter augmentation with random brightness, contrast, saturation, and hue perturbations, using the same augmentation parameters across all views in a sample to preserve multi-view consistency.
Training is performed with distributed data parallelism using a global batch size of $16$.

\paragraph{Optimization.}
We optimize the model with AdamW using learning rate $5\times10^{-4}$ and weight decay $10^{-6}$.
Gradient norms are clipped to $1.0$.
The learning-rate schedule consists of a linear warm-up followed by cosine decay.
Training runs for $220{,}000$ optimization steps.

\paragraph{Stage schedule.}
The number of exposed Gaussians per token is controlled by the stage index $s$, with
\begin{equation}
G = 2^s.
\end{equation}
We use the following schedule:
\begin{center}
\small
\begin{tabular}{c c c}
\toprule
Training step & Stage $s$ & Gaussians / token \\
\midrule
$0$--$10$k   & $0$ & $1$ \\
$10$k--$20$k & $1$ & $2$ \\
$20$k--$50$k & $2$ & $4$ \\
$>50$k       & $3$ & $8$ \\
\bottomrule
\end{tabular}
\end{center}
Transitions between stages are smoothed using linear interpolation over $2$k iterations.
\begin{table}[t]
\centering
\caption{Implementation details and training hyperparameters.}
\small
\setlength{\tabcolsep}{6pt}
\begin{tabular}{l c}
\toprule
\textbf{Parameter} & \textbf{Value} \\
\midrule
\rowcolor[gray]{0.95} \multicolumn{2}{l}{\textit{Architecture}} \\
Latent token dimension & 512 \\
Input token dimension & 512 (RGB), 256 (rays) \\
Number of latent tokens & 2048 \\
Patch size & 8 \\
Encoder blocks & 4 \\
Self-attention blocks & 2 \\
Gaussian candidates per token & 16 \\
Gaussians per token & 8 \\
Total Gaussians & 16,384 \\
\midrule
\rowcolor[gray]{0.95} \multicolumn{2}{l}{\textit{Gaussian Representation}} \\
SH degree & 3 \\
Rotation & 6D \\
Mean offset & (0, 0, 1.5) \\
Rotation offset & (1,0,0,0,1,0) \\
Log Scale offset & -2 \\
Logit Opacity offset & -5 \\
\midrule
\rowcolor[gray]{0.95} \multicolumn{2}{l}{\textit{Data \& Training}} \\
Training resolution & $256 \times 256$ \\
Target views & 12 \\
Frame distance sampling window & $[40,220]$ frames \\
Training steps & 220k \\
Global batch size & 16 \\
\midrule
\rowcolor[gray]{0.95} \multicolumn{2}{l}{\textit{Optimization \& Loss Weights}} \\
Optimizer & AdamW \\
Learning rate & $5 \times 10^{-4}$ \\
Weight decay & $10^{-6}$ \\
Gradient clipping & 1.0 \\
$\lambda_{\mathrm{mse}}$ & 2 \\
$\lambda_{\mathrm{perc}}$ & 1 \\
$\lambda_{\mathrm{fru}}$ & $10^{-2}$ \\
$\lambda_{\mathrm{dec}}$ & $10^{-2}$ \\
$\lambda_{\mathrm{con}}^{\alpha}$ & $10^{-3}$ \\
$\lambda_{\mathrm{con}}^{d}$ & $10^{-2}$ \\
\bottomrule
\end{tabular}
\label{tab:supp_impl}
\end{table}

\subsection{Training Objective and Losses}
\label{supp:losses}

In Sec. 3.5 in the main text, the training objective is written as
\begin{equation}
\mathcal{L}
=
\lambda_{\mathrm{ren}} \mathcal{L}_{\mathrm{ren}}
+
\lambda_{\mathrm{con}} \mathcal{L}_{\mathrm{con}}
+
\lambda_{\mathrm{reg}} \mathcal{L}_{\mathrm{reg}} .
\end{equation}
Here we provide the full form of all training losses used in practice.

\paragraph{Rendering loss.}
Given a target view $I_t$ and its camera $\pi_t$, we render the predicted Gaussian scene into an image $\hat{I}_t$ and optimize
\begin{equation}
\mathcal{L}_{\mathrm{ren}}
=
\lambda_{\mathrm{mse}}\|I_t - \hat{I}_t\|_2^2
+
\lambda_{\mathrm{perc}} \mathcal{L}_{\mathrm{perc}}(I_t, \hat{I}_t),
\end{equation}
where $\mathcal{L}_{\mathrm{perc}}$ is a perceptual image loss.
This is the main supervision signal.

\paragraph{Subset consistency loss.}
To encourage the model to produce compatible reconstructions from different subsets of the same scene, we first sample a temporally ordered sequence of $24$ views uniformly from the interval $[\texttt{minframe}, \texttt{maxframe}]$, where $\texttt{maxframe}-\texttt{minframe}\in[40,220]$. We then select $12$ target views uniformly from this sampled sequence, and use the remaining views as input context. The two input subsets, denoted $I^a$ and $I^b$, are constructed from the ordered context views in an interleaved manner with shared anchor views: both subsets include the first and last context views (corresponding to the minimum and maximum sampled frames), while the intermediate context views are split by parity of their order in the context sequence. Specifically, $I^a$ contains the anchor views together with every odd-indexed intermediate context view, and $I^b$ contains the same anchor views together with every even-indexed intermediate context view. This produces two overlapping subsets that share the boundary views while covering complementary temporal samples of the scene.
Both subsets are used to reconstruct the same target views.

Rather than matching latent variables directly, our implementation applies consistency in rendered space.
Let $\hat{O}^{a},\hat{O}^{b}$ denote the rendered accumulation maps, and $\hat{D}^{a},\hat{D}^{b}$ the rendered depth maps.
We use a symmetric stop-gradient formulation:
\begin{align}
\mathcal{L}_{\mathrm{con}}
&=
\lambda_{\alpha}^{\mathrm{con}} \mathcal{L}_{\mathrm{con}}^{\alpha}
+
\lambda_{d}^{\mathrm{con}} \mathcal{L}_{\mathrm{con}}^{d}, \\
\mathcal{L}_{\mathrm{con}}^{\alpha}
&=
\frac{1}{2}\|\hat{O}^{a}-\operatorname{sg}(\hat{O}^{b})\|_1
+
\frac{1}{2}\|\hat{O}^{b}-\operatorname{sg}(\hat{O}^{a})\|_1, \\
\mathcal{L}_{\mathrm{con}}^{d}
&=
\frac{1}{2}\|\hat{D}^{a}-\operatorname{sg}(\hat{D}^{b})\|_{1,\Omega}
+
\frac{1}{2}\|\hat{D}^{b}-\operatorname{sg}(\hat{D}^{a})\|_{1,\Omega}.
\end{align}
Here, $\operatorname{sg}(\cdot)$ denotes stop-gradient.
For depth consistency, $\Omega$ optionally restricts the comparison to pixels where both branches have sufficient rendered support.
This symmetric stop-gradient form provides mutual supervision between the two branches while reducing trivial co-adaptation.

\paragraph{Regularization loss.}
The regularization term is
\begin{equation}
\mathcal{L}_{\mathrm{reg}}
=
\lambda_{\mathrm{fru}}\mathcal{L}_{\mathrm{fru}}
+
\lambda_{\mathrm{dec}}\mathcal{L}_{\mathrm{dec}} .
\end{equation}
It contains a frustum constraint on Gaussian centers together with decoder-side regularization terms on Gaussian parameters.

\paragraph{Frustum constraint.}
To prevent Gaussians from drifting to unsupported regions of space, we apply a soft frustum loss to the predicted Gaussian means.
Let $\mu_n \in \mathbb{R}^3$ denote the center of Gaussian $n$.
For each input view, we transform $\mu_n$ into camera coordinates, project it to image coordinates, and measure continuous violations of the image bounds and valid depth range.
If a Gaussian lies inside at least one input-view frustum, its penalty is zero.
Otherwise, it receives a smooth positive penalty:
\begin{equation}
\mathcal{L}_{\mathrm{fru}}
=
\frac{1}{N}\sum_{n=1}^{N}
\log\left(1+\frac{v_n}{\tau}\right),
\end{equation}
where $v_n$ is the minimum frustum violation of Gaussian $n$ across the input views, and $\tau$ controls the softness of the penalty.

\paragraph{Decoder-side regularization.}
The decoder additionally returns a regularization term
\begin{equation}
\mathcal{L}_{\mathrm{dec}}
=
10^{-2}\left(
\mathcal{L}_{\mathrm{opacity}}
+
\mathcal{L}_{\mathrm{scale}}
+
\mathcal{L}_{\mathrm{rot}}
+
\mathcal{L}_{\mathrm{SH}}
\right)
\end{equation}

These terms stabilize optimization in compact Gaussian representations and reduce degenerate solutions.

\paragraph{Opacity regularization.}
We regularize opacity to prevent front splats from saturating too early.
If foreground Gaussians become highly opaque early in training, they can block gradient flow to deeper parts of the representation.
This is especially harmful in compact 3DGS settings, where a small number of Gaussians must jointly explain both visible surfaces and hidden structure.
We therefore penalize both the mean opacity and opacity logits that exceed a prescribed threshold:
\begin{equation}
\mathcal{L}_{\mathrm{opacity}}
=
\frac{1}{N}\sum_{n=1}^{N}\alpha_n
+
\frac{1}{N}\sum_{n=1}^{N}\max(o_n - t,0)^2,
\end{equation}
where $\alpha_n = \sigma(o_n)$ is the opacity of Gaussian $n$, $o_n$ is its opacity logit, and
\begin{equation}
t = \log\frac{\alpha_{\max}}{1-\alpha_{\max}}
\end{equation}
is the logit corresponding to a target maximum opacity $\alpha_{\max}$.
In our implementation, $\alpha_{\max}=0.2$.

\paragraph{Scale regularization.}
Gaussian scales are predicted in log-space.
Before clamping, we penalize scales that exceed a predefined maximum:
\begin{equation}
\mathcal{L}_{\mathrm{scale}}
=
\frac{1}{N}\sum_{n=1}^{N}
\max(\log s_n - \log s_{\max},0)^2.
\end{equation}

\paragraph{Rotation regularization.}
The decoder predicts residual 6D rotations.
We apply a small quadratic penalty to these raw residuals:
\begin{equation}
\mathcal{L}_{\mathrm{rot}} = \frac{1}{N}\sum_{n=1}^{N}\|\mathbf{r}_n\|_2^2.
\end{equation}

\paragraph{SH soft-cap regularization.}
To avoid unstable appearance coefficients, we softly penalize spherical harmonics coefficients whose magnitude exceeds a prescribed cap:
\begin{equation}
\mathcal{L}_{\mathrm{SH}}
=
\mathbb{E}\left[
\left(
\operatorname{softplus}\left(\frac{|c|-c_{\max}}{\tau_{\mathrm{SH}}}\right)\tau_{\mathrm{SH}}
\right)^p
\right].
\end{equation}

\paragraph{Final objective.}
When subset consistency is enabled, we perform two forward passes, one for each input subset, and compute supervised rendering losses for both.
The final objective becomes
\begin{equation}
\mathcal{L}
=
\frac{1}{2}\left(
\mathcal{L}_{\mathrm{subset}}^{a}
+
\mathcal{L}_{\mathrm{subset}}^{b}
\right)
+
\mathcal{L}_{\mathrm{con}},
\end{equation}
where each term is
\begin{equation}
\mathcal{L}_{\mathrm{subset}}^{k}
=
\mathcal{L}_{\mathrm{ren}}^{k}
+
\mathcal{L}_{\mathrm{reg}}^{k},
\qquad k \in \{a,b\}.
\end{equation}
When subset consistency is disabled, training falls back to the standard single-branch supervised objective.

\noindent\textit{All loss weights are provided in Tab.~\ref{tab:supp_impl}.
}
\section{Additional Evalution Details}
\label{sec:supp_baseline_details}

\subsection{Baselines Evaluation Details}

\paragraph{DepthSplat.}
We use the large DepthSplat model under our $256\times256$ RealEstate10K evaluation protocol using the official repository and weights. 
\paragraph{ZPressor.}
 The reported ZPressor result in the main text uses \textbf{6 anchor views}. We additionally report a more aggressive \textbf{3-anchor} variant in Tab.~\ref{tab:main_results}. 
We use the MVSplat+Zpressor variant because it is the reported best model trained only on RealEstate10K, 
using the official repository and weights. 
\paragraph{GGN.}
We use the official public implementation and checkpoints. 
In our setup, we report the settings for which the public code produced valid reconstructions. In particular, we use the official pipeline for the 12- and 24-view settings, whereas the publicly available implementation did not yield stable reconstructions for our 36-view evaluation and is therefore omitted.

\paragraph{C3G.}
We use the released \texttt{gaussian\_decoder\_multiview.ckpt}. The official C3G codebase provides both a 2-view Gaussian decoder and a dedicated multiview Gaussian decoder, together with multiview RealEstate10K evaluation code. The codebase also explicitly states that it builds on VGGT and NoPoSplat. We further state that the C3G setup includes a post-inference pose optimization. 

\paragraph{AnySplat.}
The numbers reported in our tables are taken from the results published by C3G. We opted for this because the official AnySplat release does not match our experimental setup.

\paragraph{NoPoSplat.}
The numbers in our tables are taken directly from the C3G benchmark. We rely on this baseline because the official NoPoSplat release focuses on sparse, unposed settings, providing checkpoints primarily for 2-view inputs and a 3-view RE10K variant. In contrast, the C3G evaluation framework adapts VGGT and NoPoSplat components to support the multi-view setting required for our comparison.

\paragraph{EcoSplat.} We report the comparison numbers provided in the original paper rather than performing a fully matched rerun within our own pipeline.

\paragraph{LVSM.}

We use the official implementation and evaluation protocol using the decoder only $256\times256$ model.

\subsection{Efficiency Benchmark Protocol}
\label{sec:supp_efficiency}

All efficiency benchmarks were measured on a single NVIDIA A100 GPU with 64\,GB VRAM, using 8 CPU cores and 128\,GB system RAM. We use the benchmarking utility shipped with each model, following the standard benchmark implementation inherited from the PixelSplat codebase \cite{charatan2024pixelsplat}. All methods are evaluated on the full test set under their respective evaluation pipelines.

For reconstruction time, we discard the first measured sample and report timings over the remaining test-set runs, in order to avoid initialization and warm-up effects. We report reconstruction time as the time required to predict the scene representation from the input views, and peak GPU memory as recorded during the same evaluation process.

%% file: figures/acid_visual.tex
\begin{figure*}[h!]
    \centering

    \newcommand{\imgwidth}{0.16\linewidth}

    \setlength{\tabcolsep}{1pt}
    \renewcommand{\arraystretch}{0.5}

    \begin{tabular}{cccccc}
        \small Zpressor & \small DepthSplat & \small GGN & \small C3G & \small Ours & \small GT \\
        \vspace{1pt} \\

        \includegraphics[width=\imgwidth]{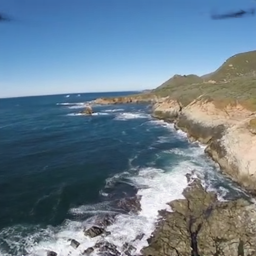} &
        \includegraphics[width=\imgwidth]{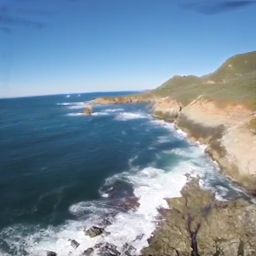} &
        \includegraphics[width=\imgwidth]{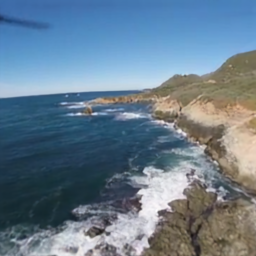} &
        \includegraphics[width=\imgwidth]{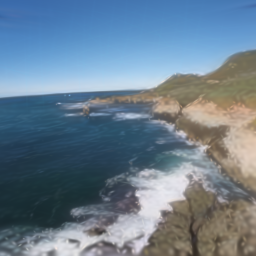} &
        \includegraphics[width=\imgwidth]{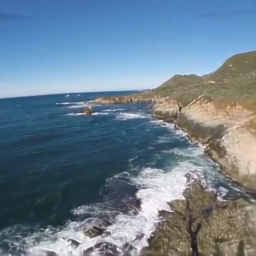} &
        \includegraphics[width=\imgwidth]{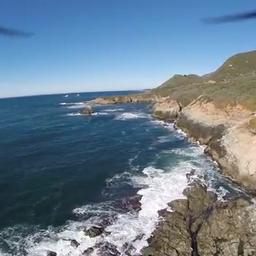} \\

        \includegraphics[width=\imgwidth]{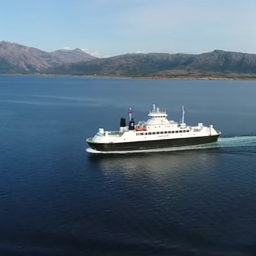} &
        \includegraphics[width=\imgwidth]{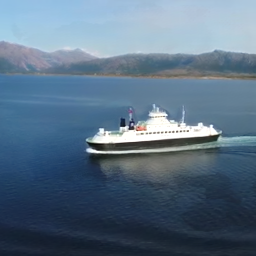} &
        \includegraphics[width=\imgwidth]{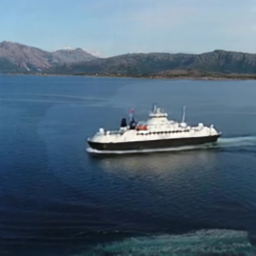} &
        \includegraphics[width=\imgwidth]{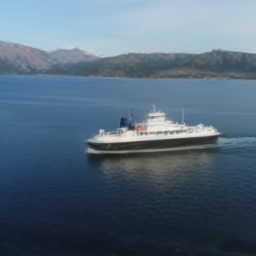} &
        \includegraphics[width=\imgwidth]{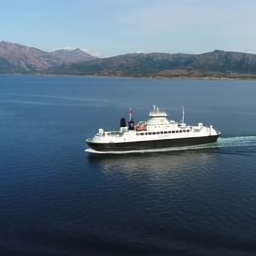} &
        \includegraphics[width=\imgwidth]{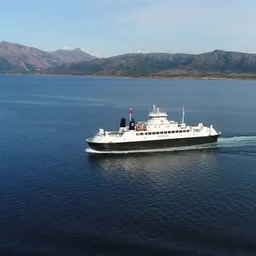} \\

        \includegraphics[width=\imgwidth]{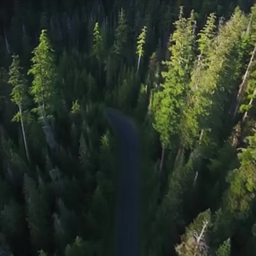} &
        \includegraphics[width=\imgwidth]{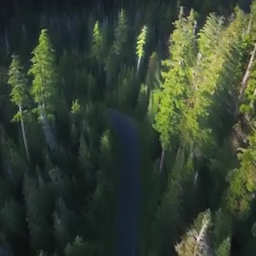} &
        \includegraphics[width=\imgwidth]{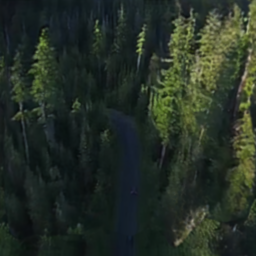} &
        \includegraphics[width=\imgwidth]{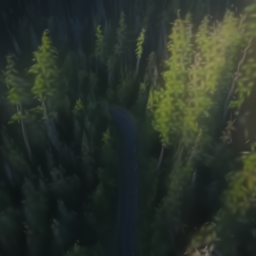} &
        \includegraphics[width=\imgwidth]{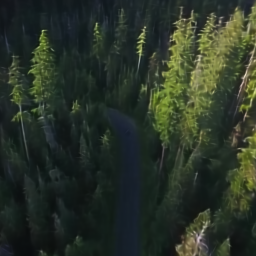} &
        \includegraphics[width=\imgwidth]{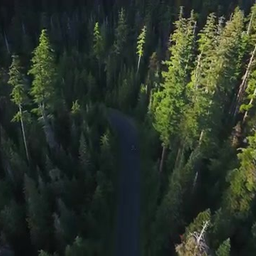} \\

        \includegraphics[width=\imgwidth]{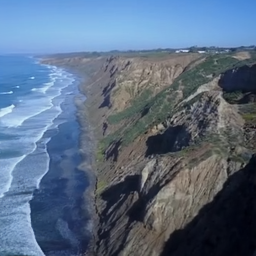} &
        \includegraphics[width=\imgwidth]{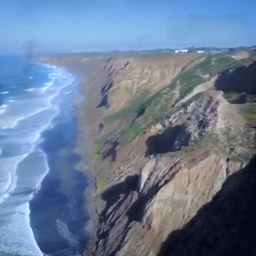} &
        \includegraphics[width=\imgwidth]{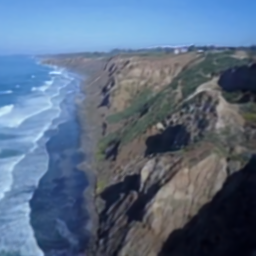} &
        \includegraphics[width=\imgwidth]{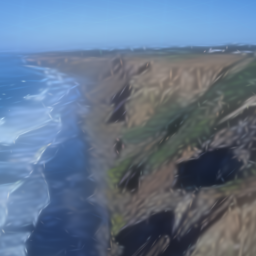} &
        \includegraphics[width=\imgwidth]{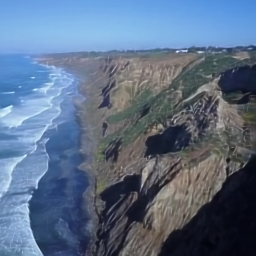} &
        \includegraphics[width=\imgwidth]{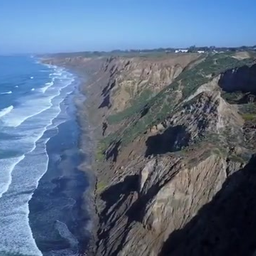} \\

        \includegraphics[width=\imgwidth]{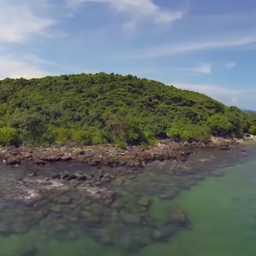} &
        \includegraphics[width=\imgwidth]{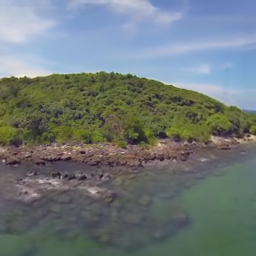} &
        \includegraphics[width=\imgwidth]{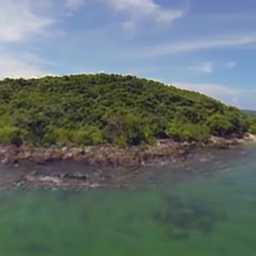} &
        \includegraphics[width=\imgwidth]{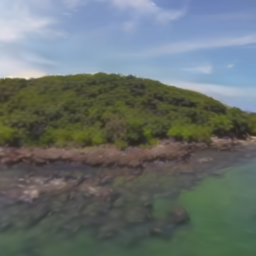} &
        \includegraphics[width=\imgwidth]{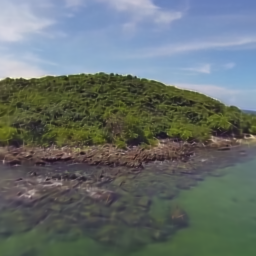} &
        \includegraphics[width=\imgwidth]{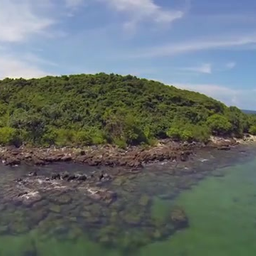} \\

        \includegraphics[width=\imgwidth]{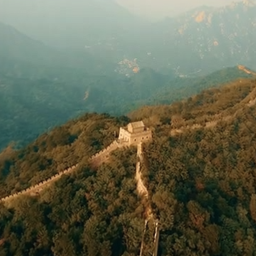} &
        \includegraphics[width=\imgwidth]{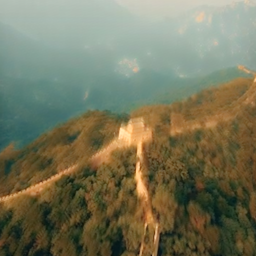} &
        \includegraphics[width=\imgwidth]{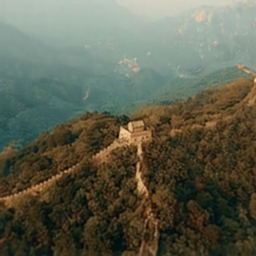} &
        \includegraphics[width=\imgwidth]{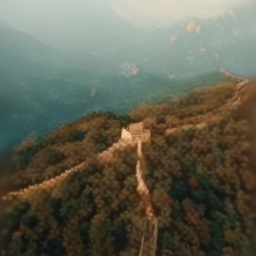} &
        \includegraphics[width=\imgwidth]{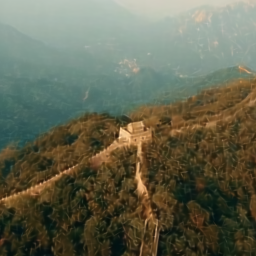} &
        \includegraphics[width=\imgwidth]{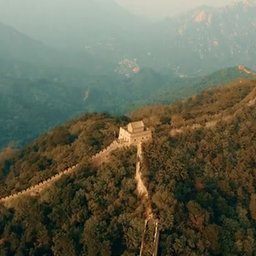} \\
    \end{tabular}

    \caption{\textbf{Qualitative comparison on ACID}. We compare \methodname{} against baselines (Zpressor, DepthSplat, GGN, C3G) and the ground truth (GT) across 6 different ACID scenes (rows).}
    \label{fig:qualitative_comparison_acid}
    \vspace{-0.5cm}
\end{figure*}